\documentclass[journal]{./IEEEtran/IEEEtran}

\usepackage{amsmath} 
\usepackage{amsfonts}
\usepackage{mathtools} 
\usepackage{amssymb}
\usepackage{graphicx}
\usepackage{bm}  
\usepackage[colorinlistoftodos, textsize=footnotesize]{todonotes}
\usepackage[colorlinks=true, allcolors=black]{hyperref}
\usepackage{setspace}
\usepackage{algorithm}      
\usepackage{algpseudocode}  
\usepackage{tikz}
\usepackage{verbatim}
\usepackage{xcolor}
\usepackage{subcaption}
\usepackage{svg}

\usepackage{adjustbox}

\usepackage[normalem]{ulem}

\usepackage[thinc]{esdiff}

\usepackage{amsthm}

\usepackage{mathrsfs}

\newif\ifcompile

\DeclareMathOperator*{\argmax}{arg\,max}

\newcommand{\reviewquestion}[2]{#2}

\newcommand{\vf}[1]{{\bm{#1}}}
\newcommand{\mf}[1]{{\mathbf{#1}}}
\newcommand{\bgamma}{{\bm\gamma}} 

\newcommand{\bxi}{{\bm\xi}}
\newcommand{\btheta}{{\bm\theta}}

\usepackage{threeparttable} 

\begin{document}

\title{Irrotational Contact Fields}


\author{Alejandro Castro, Xuchen Han, Joseph Masterjohn
\thanks{All the authors are with the Toyota Research Institute, USA, {\tt\small
firstname.lastname@tri.global}.}%
}


\maketitle

\begin{abstract}
We present a framework for generating convex approximations of complex contact
models, incorporating experimentally validated models like Hunt \& Crossley
coupled with Coulomb's law of friction alongside the principle of maximum
dissipation. Our approach is robust across a wide range of stiffness values,
making it suitable for both compliant surfaces and rigid approximations. We
evaluate these approximations across a wide variety of test cases, detailing
properties and limitations. We implement a fully differentiable solution in the
open-source robotics toolkit, Drake. Our novel \emph{hybrid} approach enables
efficient computation of gradients for complex geometric models by reusing
factorizations from contact resolution. We demonstrate robust simulation of
robotic tasks at interactive rates, with accurately resolved stiction and
contact transitions, supporting effective sim-to-real transfer.

\end{abstract}


%
\IEEEpeerreviewmaketitle

\section{Introduction}

\IEEEPARstart{S}{imulation} of multibody systems with frictional contact is
essential in robotics, aiding hardware optimization, controller design and
testing, continuous integration, and data generation for machine learning.
Accurate physics models enable advanced controller design and trajectory
optimization algorithms. However, achieving robust, accurate, and efficient
simulations for contact-rich robotics applications remains challenging.

Rigid body dynamics with frictional contact are complicated by non-smooth
solutions. Acceleration-level formulations can lead to singular configurations,
known as Painlev\'e paradoxes, where solutions may not exist. Discrete
velocity-level formulations circumvent this by allowing discrete velocity jumps and
impulsive forces. While many variants exist, the general form of a discrete
formulation enforces balance of momentum subject to contact constraints,
with additional constraints to incorporate Coulomb's law of friction under the
maximum dissipation principle. With the addition of decision
variables and Lagrange multipliers, the result is a large and challenging-to-solve
Non-linear Complementarity Problem (NCP) with a much larger number of
variables than the original problem.

Solving NCPs robustly and efficiently has remained elusive. NCPs are equivalent
to non-convex global optimization problems, which are generally
NP-hard~\cite{bib:Kaufman2008}. Therefore, NCPs may lack solutions or have
multiple solutions. In practice, iterative solvers might incorrectly assume
convergence or terminate early, leading to solutions that fail to satisfy the
original equations or violate physical laws.

There are other, less frequently-discussed issues, such as numerical conditioning
and implementation details, that affect convergence properties and robustness. These
are important problems when it comes to the simulation of complex systems, with
either many degrees of freedom (DOFs), a large number of constraints,
or both. Solutions that work for small systems often do not work when faced
with real-world engineering applications.

In an attempt to make the problem tractable, Anitescu introduced a convex
approximation of contact~\cite{bib:anitescu2006}, effectively replacing
the original NCP with a convex approximation that guarantees the existence of solutions.
Later, Todorov~\cite{bib:todorov2011} regularized this formulation to write a strictly
convex formulation with a unique solution. Our previous work~\cite{bib:castro2022unconstrained}
introduced the convex Semi-Analytic Primal
(SAP) formulation for modeling compliant contact. SAP embraces compliance
to provide a robust and performant tool that targets robotics applications ---
where grippers feature compliant surfaces for stable grasps and robotic feet
are often padded with compliant materials. Although the formulation is
inherently compliant, rigid contact can be modeled to a good approximation using a
\emph{near-rigid} approximation. The work focused on physical correctness,
numerical conditioning, and robustness.

However, the SAP formulation has several limitations. The most well-known is
the artifact of \emph{gliding} during slip, an artifact inherited from the
original formulation by Anitescu~\cite{bib:mazhar2014} and shared by Todorov's
formulation~\cite{bib:todorov2011} in MuJoCo \cite{bib:todorov2012mujoco}. In
Anitescu's formulation, as objects slip, they glide at a finite distance $\delta
t\mu\Vert\vf{v}_t\Vert$, proportional to the time step $\delta t$, the
coefficient of friction $\mu$, and the slip speed $\Vert\vf{v}_t\Vert$.
\reviewquestion{Q12}{For interactive simulation of a pushing task at
$\Vert\vf{v}_t\Vert\approx 1 \text{ m/s}$ with $\delta t=10\text{ ms}$ and
$\mu=1$, the gliding artifact is as large as $1\text{ cm}$. Similarly, this
gliding artifact was reported inadequate for the simulation of quadrupeds in
\cite{bib:lelidec2024}. Moreover, we show that both SAP's compliant model and
Todorov's regularized model are inconsistent. The gliding effect is also
proportional to dissipation and does not vanish as the time step goes to zero.
This exacerbates the nonphysical gliding artifact further, potentially
increasing action at a distance orders of magnitude depending on the amount of
dissipation, even at small time step sizes.} Finally, SAP's model of compliance
is intrinsic to its convex formulation, and therefore, it is not possible to
incorporate experimentally validated engineering-grade models of contact, such
as Hertz and Hunt \& Crossley~\cite{bib:hunt_crossley}. Prior to this work, no
convex formulation allowed the integration of such models.

\reviewquestion{Q1, Q2, Q17, Q21}{Despite its limitations, the SAP formulation
offers a robust solution with theoretical convergence guarantees that carry over
to practical implementations. This work develops novel convex approximations
that retain these guarantees while reducing (and even eliminating)
\emph{gliding} and related artifacts of existing convex formulations, enabling
the use of validated contact models, enhancing friction fidelity, and improving
numerical performance.}

\reviewquestion{Q15, Q22}{Our work is organized as follows. Section
\ref{sec:optimization_framework} summarizes our previous work and reviews
compliant contact with regularized friction. Section
\ref{sec:irrotational_contact_fields} presents the Irrotational Contact Fields
framework, establishing conditions for embedding arbitrary contact models into
our convex formulation. Section \ref{sec:models_family} derives two new
approximations, \emph{Lagged} and \emph{Similar}, by solving these conditions.
Section \ref{sec:convex_hunt_crossley} completes the model using Hunt \&
Crossley dissipation, enabling, for the first time, integration of arbitrary
contact laws in a convex setting. We analyze the approximations in Section
\ref{sec:comparative_analysis}, succinctly summarizing their properties in Table
\ref{tab:models_properties}. Section \ref{sec:impacts_and_conditioning} analyzes
the connection between impacts and numerical conditioning (we believe this is
the first time this connection has been made) and Section
\ref{sec:barriers_and_compliance} explores the connection to barrier methods.
Implementation details appear in Section \ref{sec:implementation}, followed by
extensive benchmarks and applications in Sections
\ref{sec:results}-\ref{sec:applications}. We close with final remarks and
conclusions in Section \ref{sec:conclusions}}

\section{\reviewquestion{Q2, Q4, Q17, Q21}{Novel Contributions}}

Our main contribution is a novel mathematical framework for generating convex
approximations of frictional contact, centered on the concept of
\emph{irrotational contact fields}.

While our previous work \cite{bib:castro2022unconstrained} draws an explicit
connection between earlier convex formulations~\cite{bib:anitescu2006,
bib:todorov2011} and the physics of compliance with regularized friction,
\reviewquestion{Q2}{irrotational contact fields enable the integration of
engineering-grade, experimentally validated, contact laws; this is something not
possible with existing convex models. Moreover, the formulation imposes minimum
requirements on the functional form used for the regularization of friction,
allowing for the design of regularized friction models with better numerical
properties.}

Within this framework, we propose two new contact approximations, \emph{Lagged}
and \emph{Similar}. Both approximations allow for the incorporation of arbitrary contact
force laws. However, we show that only the Lagged approximation eliminates the
\emph{gliding} at a distance artifact of SAP \cite{bib:castro2022unconstrained}
and MuJoCo \cite{bib:todorov2012mujoco} formulations.

Our key contributions are:
\begin{enumerate}
    \item A mathematical framework to generate convex approximations of complex
    contact models.
    \item The first convex approximation of compliant contact with Hunt \&
    Crossley dissipation \cite{bib:hunt_crossley}.
    \item Two new convex approximations, Lagged and Similar.
\end{enumerate}

\reviewquestion{Q2, Q4}{In a nutshell, our \emph{Irrotational Contact Fields}
framework addresses two key challenges that are impossible to resolve with existing
convex formulations:
\begin{enumerate}
    \item To incorporate arbitrary engineering-grade models of contact (e.g.
    Hunt \& Crossley).
    \item To eliminate the \emph{gliding} artifact of existing convex
    formulations. \end{enumerate}}

\reviewquestion{Q4}{In addition to these main research contributions, our work
includes a fully differentiable open-source implementation in
Drake~\cite{bib:drake}; see Section \ref{sec:gradients}. We compare our convex
approximations in Section~\ref{sec:comparative_analysis}, validate them through
extensive simulation tests in Section~\ref{sec:results}, and stress test them in
Section~\ref{sec:applications} with engineering applications, including the
complex gearing mechanism of a BarrettHand~\cite{bib:barretthand} (Section
\ref{sec:barrett_hand}) and deformable FEM bodies (Section
\ref{sec:deformable}). Moreover, we provide an extensive study on a Franka hand
to assess the impact of different approximations on grasp stability (Section
\ref{sec:franka_hand}).}

\section{Mathematical Formulation}
\label{sec:optimization_framework}

\subsection{Preliminaries}

We first introduce the notation and conventions used in our framework. We
closely follow the notation in our previous
work~\cite{bib:castro2022unconstrained,bib:masterjohn2021discrete}, later
extended to deformable Finite Element Method (FEM) models in~\cite{bib:han2023}
and Material Point Method (MPM) models in~\cite{bib:zong2024_convex_mpm}.

State is described by generalized positions $\mf{q} \in \mathbb{R}^{n_q}$ and
generalized velocities $\mf{v} \in \mathbb{R}^{n_v}$, where $n_q$ and $n_v$ are the
number of generalized positions and velocities, respectively. Time derivatives
of the generalized positions relate to the generalized velocities by the kinematic map
$\dot{\mf{q}} = \mf{N}(\mf{q})\mf{v}$. We use joint coordinates to describe
articulated rigid bodies.

Consider a system with $n_c$ constraints introducing $n_\text{eq}$ constraint
equations. Constraint velocities $\mf{v}_c\in\mathbb{R}^{n_\text{eq}}$ are given
by $\mf{v}_c=\mf{J}\mf{v}+\mf{b}$, where $\mf{J}$ is the stacked Jacobian for
all constraints and $\mf{b}$ is a bias term. Following
\cite{bib:castro2022unconstrained}, we write a symplectic Euler scheme with
fixed time steps of size $\delta t$ as
\begin{flalign}
	&\mf{M}(\mf{q}_0)(\mf{v}-\mf{v}_0) =\nonumber\\
    &\qquad\delta t\,\mf{k}_1(\mf{q},\mf{v}) + \delta t\,\mf{k}_2(\mf{q}_0,\mf{v}_0) +
	\mf{J}(\mf{q}_0)^T\mf{\bgamma},
    \label{eq:scheme_momentum}\\
    &\mf{q} = \mf{q}_0 + \delta t \mf{N}(\mf{q}_0)\mf{v}\nonumber,
\end{flalign}
where quantities with the naught subscript are evaluated at the previous time
step $t^n$ and quantities without subscripts are evaluated at the next time step
$t^{n+1}= t^n+\delta t$. $\mf{M}$ is the joint space mass matrix. Stiff terms,
such as springs, and stabilizing terms, such as damping and rotor inertias, are
treated implicitly in $\mf{k}_1$. Smooth, non-stiff terms such as gravity,
gyroscopic forces, and feed-forward actuation are treated explicitly in
$\mf{k}_2$. Vector $\bgamma\in\mathbb{R}^{n_\text{eq}}$ corresponds to
constraint impulses. Our implementation in Drake~\cite{bib:drake} includes
contact constraints, holonomic constraints (e.g. weld, distance, coupler), and PD
controllers with effort limits \cite{bib:castro2022unconstrained, bib:han2023}.

We define the momentum residual $\mf{m}(\mf{v})$ as
\begin{equation*}
    \mf{m}(\mf{v}) = \mf{M}(\mf{q}_0)(\mf{v}-\mf{v}_0) -
    \delta t\,\mf{k}_1(\mf{q},\mf{v}) - \delta t\,\mf{k}_2(\mf{q}_0,\mf{v}_0),
\end{equation*}
and define the \emph{free motion velocities} $\mf{v}^*$ as the solution to
$\mf{m}(\mf{v}^*)=\mf{0}$, that is, the velocities the system would have in the
absence of constraint forces. SAP \cite{bib:castro2022unconstrained} linearizes
\eqref{eq:scheme_momentum} at $\mf{v} = \mf{v}^*$ as
\begin{eqnarray}
	\mf{A}(\mf{v}-\mf{v}^*) = \mf{J}^T\bgamma,
    \label{eq:optimality_condition}
\end{eqnarray}
where $\mf{A}$ is a symmetric positive-definite (SPD) approximation of the gradient of the momentum residual
accurate to first order, i.e.
\begin{align*}
	\left. \frac{\partial \mf{m}}{\partial \mf{v}} \right|_{\mf{v}=\mf{v}^*} = \mf{A} + \mathcal{O}(\delta t).
\end{align*}

While early work~\cite{bib:stewart1996implicit, bib:anitescu1997} use $\mf{A} =
\mf{M}(\mf{q}_0)$, the mass matrix at the previous time step, our formulation
incorporates the modeling of joint springs, damping, and rotor inertia
implicitly~\cite{bib:castro2022unconstrained}. 

Traditional NCP formulations supplement~\eqref{eq:optimality_condition} with
constraint equations to model contact. Additional constraint equations, slack
variables, and Lagrange multipliers lead to a large and challenging-to-solve NCP.
Instead, SAP~\cite{bib:castro2022unconstrained} follows a different approach.
Inspired by the analytical inverse dynamics from Todorov~\cite{bib:todorov2011},
SAP eliminates constraints analytically to write an unconstrained convex
optimization problem for the velocities at the next time step
\begin{eqnarray}
	\min_{\mf{v}} \ell_p(\mf{v}) = \frac{1}{2}\Vert\mf{v}-\mf{v}^*\Vert_{A}^2 +
	\ell_c(\mf{v}),
    \label{eq:unconstrained_fomulation}
\end{eqnarray}
where $\Vert\mf{x}\Vert_{A}^2 = \mf{x}^T\mf{A}\,\mf{x}$, and the constraints
cost $\ell_c(\mf{v})$ penalizes constraint impulses. It can be shown that
SAP's~\cite{bib:castro2022unconstrained} formulation models linear compliant
contact with a Kelvin-Voigt (linear) model of damping. The formulation is
strongly convex, with convergence guarantees that lead to robust software
implementations. Although the formulation is inherently compliant with
regularized friction, \cite{bib:castro2022unconstrained}~shows how rigid contact
can be modeled to a good approximation using a \emph{near-rigid} approximation.


To be more precise, we write $\ell_c(\mf{v}; \mf{x}_0)$, emphasizing this is an
\emph{incremental potential} \cite{bib:pandolfi2002}. This potential depends on
the generalized velocities $\mf{v}$, \emph{given} a previous state $\mf{x}_0$,
making it inherently discrete---unlike continuous potentials like gravity or
electric fields. The presence of the quadratic term $\Vert\mf{x}\Vert_{A}^2$
ensures strong convexity of the optimization problem
\eqref{eq:unconstrained_fomulation} when $\ell_c$ is convex, guaranteeing a
unique solution. Developing such convex potentials for frictional contact
modeling is the primary aim of this work.


Taking the gradient of~\eqref{eq:unconstrained_fomulation}, we obtain the balance
of momentum~\eqref{eq:optimality_condition} and find that impulses emerge as
a result of the constraint potentials
\begin{eqnarray}
    \bgamma(\mf{v}_c; \mf{x}_0) = -\frac{\partial\ell_c(\mf{v}_c;
    \mf{x}_0)}{\partial\mf{v}_c},
    \label{eq:gamma_from_potential}
\end{eqnarray}
where we use the notation $\partial f/\partial\vf{x} = \nabla f \in\mathbb{R}^n$
to denote the gradient of a scalar function $f(\vf{x}):
\mathbb{R}^n\rightarrow\mathbb{R}$. As with SAP, we consider potential functions
that are
\emph{separable}
\begin{eqnarray}
    \ell_c(\mf{v}_c; \mf{x}_0) = \sum_{i=1}^{n_c}\ell_{c,i}(\vf{v}_{c,i};
    \mf{x}_0),
    \label{eq:separable_potential}
\end{eqnarray}
where $\mf{v}_c$ is the stacked vector of individual constraint velocities
$\vf{v}_{c,i}$. For contact constraints, we define a contact frame $C_i$ for
which we arbitrarily choose the $z\text{-axis}$ to coincide with the contact
normal $\hat{\vf{n}}_i$. In this frame, the normal and tangential components of
$\vf{v}_{c,i}$ are given by $v_{n,i} = \hat{\vf{n}}_i\cdot\vf{v}_{c,i}$ and
$\vf{v}_{t,i} = \vf{v}_{c,i}-v_{n,i}\hat{\vf{n}}_i$ respectively, and
$\vf{v}_{c,i}=[\vf{v}_{t,i}\,v_{n,i}]$. By convention, we define the relative
contact velocity $\vf{v}_{c,i}$ and normal $\hat{\vf{n}}_i$ such that $v_{n,i} >
0$ for objects moving away from each other.

Using this separable potential \eqref{eq:separable_potential}
in~\eqref{eq:gamma_from_potential}, the impulse vector $\bgamma$ is the stacked
vector of individual constraint contributions $\bgamma_i(\mf{v}_{c,i}; \mf{x}_0)
= -\partial\ell_{c,i}/\partial\mf{v}_{c,i}$. We highlight the dependence
on the previous time step state $\mf{x}_0$ to emphasize the discrete (or
incremental) nature of these potentials and the resulting impulses. However,
hereinafter, we write $\bgamma_i(\vf{v}_{c,i})$ and $\ell_{c,i}(\vf{v}_{c,i})$ to
shorten notation, and the functional dependence on the previous time step state
is implicitly assumed.

\subsection{Compliant Contact with Regularized Friction}
\label{sec:compliant_contact}

Compliant contact models are widely adopted in the literature for engineering
applications \cite{bib:flores2021contact}. Experimentally validated models based
on Hertz theory \cite{bib:johnson1987_contact_mechanics} with Hunt \& Crossley
dissipation \cite{bib:hunt_crossley} are common examples.

In this work, we consider a general force law 
\begin{equation}
    f_n(\phi, v_n),
    \label{eq:generic_force_law}
\end{equation}
function of the signed distance $\phi$ (defined as negative when objects overlap),
and of the normal velocity $v_n$ (defined positive when objects separate). In
the discrete setting, with a time step of size $\delta t$, we use a first-order
approximation of the signed distance $\phi=\phi_0+\delta t v_n$, implicit in the
next time step velocity $v_n$. Using this approximation, we define a discrete
normal impulse as
\begin{equation}
    n(v_n; \phi_0) = \delta t\,f_n(\phi_0+\delta t\,v_n, v_n),
    \label{eq:discrete_normal_impulse}
\end{equation}
and the associated normal potential $\ell_n(v_n)=-N(v_n)$, with $N(v_n)$ the
indefinite integral of $n(v_n)$. That is, $\gamma_n = -\ell_n'(v_n)=N'(v_n)$. We
observe that since
\begin{equation*}
    \frac{d^2\ell_n}{dv_n^2} = -\delta t^2\frac{\partial f_n}{\partial\phi} - \delta t\frac{\partial f_n}{\partial v_n},
\end{equation*}
it follows that
\begin{equation}
    \frac{\partial f_n}{\partial\phi}\le 0\text{ and }
    \frac{\partial f_n}{\partial v_n}\le 0,
    \label{eq:normal_sufficient_conditions}
\end{equation}
are sufficient conditions for the normal cost $\ell_n(v_n)$ to be convex.

Friction can be modeled as a continuous function of state using a
\emph{regularized} approximation
\begin{eqnarray}
    \bgamma_t(\vf{v}_c) &=&
    -\mu\,f(\Vert\vf{v}_t\Vert/\varepsilon_s)\,n(v_n)\,\hat{\vf{t}}\nonumber\\
    f(s) &=& \frac{s}{\sqrt{1+s^2}}
    \label{eq:regularized_friction}\\
    \hat{\vf{t}} &=& \frac{\vf{v}_t}{\Vert\vf{v}_t\Vert}\nonumber
\end{eqnarray}
where function $f(s)\le 1$ \emph{regularizes} Coulomb friction, with $\mu$ the
coefficient of friction and $\varepsilon_s$ the \emph{regularization parameter}.
When $\Vert\vf{v}_t\Vert\ll\varepsilon_s$, the model behaves as viscous damping
with high viscosity. When $\Vert\vf{v}_t\Vert \gg \varepsilon_s$, the model
approximates Coulomb's law, with friction opposing slip velocity according to
the maximum dissipation principle and $\Vert\bgamma_t\Vert \rightarrow
\mu\gamma_n$. The choice of function $f(s)$ is somewhat arbitrary as long as
$f(s) \le 1$ and $f(s)=0$ at $s=0$. We choose $f(s)$ such that
\eqref{eq:regularized_friction}~can be simplified to
\begin{eqnarray}
    \bgamma_t(\vf{v}_c) &=& -\mu\,n(v_n)\,\hat{\vf{t}}_s\nonumber\\
    \hat{\vf{t}}_s &=&
    \frac{\vf{v}_t}{\sqrt{\Vert\vf{v}_t\Vert^2+\varepsilon_s^2}}
    \label{eq:regularized_friction_soft}
\end{eqnarray}
where we define the regularized or \emph{soft} tangent vector $\hat{\vf{t}}_s$,
which can be shown to be the gradient with respect to $\vf{v}_t$ of the
\emph{soft} norm $\Vert\vf{x}\Vert_s=\sqrt{\Vert\vf{x}\Vert^2+\varepsilon_s^2} -
\varepsilon_s$ (Appendix~\ref{sec:soft_norms}). Unlike $\hat{\vf{t}}$, which is
not well-defined at $\vf{v}_t=\vf{0}$, $\hat{\vf{t}}_s$ is well-defined and
continuous for all values of slip velocity. Moreover,
\eqref{eq:regularized_friction_soft} has continuous gradients, a  desirable
property that improves the convergence of non-linear solvers based on Newton's
method.

In total, we write the contact impulse as
\begin{eqnarray}
	\bgamma(\vf{v}_c) &=&
    \begin{bmatrix}
		\bgamma_t\\
		\gamma_n \end{bmatrix}=
	\begin{bmatrix}
		-\mu\,\hat{\vf{t}}_s(\vf{v}_t)\,n(v_n)\\
		n(v_n) \end{bmatrix}.
    \label{eq:total_impulse}
\end{eqnarray}

However, this model is not necessarily the result of a potential and
$\bgamma(\vf{v}_{c}) \ne -\partial\ell_{c}/\partial\vf{v}_{c}$ in general. In
the next section we develop a general theory that allows us to write convex
approximations of this model that fit within the optimization framework
\eqref{eq:unconstrained_fomulation}.

\section{Irrotational Contact Fields}
\label{sec:irrotational_contact_fields}

Helmholtz's theorem~\cite{bib:bhatia2012} states that any vector field admits
a decomposition into an irrotational field (zero curl) and a solenoidal field
(zero divergence)
\begin{eqnarray}
    \bgamma(\vf{v}_c) = -\frac{\partial\ell(\vf{v}_c)}{\partial\vf{v}_c} +
    \nabla\times\vf{A}(\vf{v}_c)
    \label{eq:helmholtz_decomposition}
\end{eqnarray}
with $\ell(\vf{v}_c)$ a scalar potential and $\vf{A}(\vf{v}_c)$ a \emph{vector
potential}.

Here, we neglect the solenoidal component to investigate \emph{irrotational
fields}, which satisfy the condition
\begin{equation}
    \nabla\times\bgamma = \mf{0}.
    \label{eq:gamma_curl}
\end{equation}

The normal component in~\eqref{eq:gamma_curl}
\begin{equation}
    \frac{\partial\gamma_{t,1}}{\partial v_{t,2}}=\frac{\partial\gamma_{t,2}}{\partial v_{t,1}},
    \label{eq:curl_normal}
\end{equation}
states that the two-dimensional field $\bgamma_t(\vf{v}_t)$ is irrotational in the
$\vf{v}_t$ plane. Consider the generic isotropic friction model
\begin{equation}
    \bgamma_t=g(\Vert\mf{v}_t\Vert, v_n)\hat{\mf{t}}
    \label{eq:generic_friction_model}
\end{equation}
whose gradient is
\begin{equation*}
    \frac{\partial\bgamma_t}{\partial\mf{v}_t}=\frac{\partial g}{\partial \Vert\mf{v}_t\Vert}\mf{P}+g\frac{\mf{P}^\perp}{\Vert\mf{v}_t\Vert}
\end{equation*}
with symmetric projection matrices $\mf{P}$ and $\mf{P}^\perp$ (see
Appendix~\ref{sec:soft_norms}). Therefore, $\partial\bgamma_t/\partial\mf{v}_t$
is symmetric, and condition~\eqref{eq:curl_normal} is satisfied. Thus isotropic
friction fields are irrotational. While this work focuses on isotropic friction,
Appendix \ref{sec:anisotropic_friction} shows how to incorporate anisotropic
friction within this same framework.

Finally, the tangential components in~\eqref{eq:gamma_curl} lead to the
condition
\begin{equation}
    \frac{\partial\bgamma_t}{\partial v_n}=\frac{\partial\gamma_n}{\partial\mf{v}_t},
    \label{eq:curl_tangent}
\end{equation}

Contact models are not irrotational in the general case, and condition
\eqref{eq:curl_tangent} is not satisfied. In the next section we solve for
irrotational approximations of \eqref{eq:total_impulse} that satisfy the curl
condition $\nabla\times\bgamma = \mf{0}$ and thus a scalar potential
$\ell(\vf{v}_c)$ exists.

\section{A Family of Convex Approximations}
\label{sec:models_family}

We present a family of convex approximations of the model in
 \eqref{eq:total_impulse} satisfying \eqref{eq:curl_tangent}, and establish
 conditions for convexity. Moreover, we present two novel convex approximations
 of contact: \emph{Lagged} and \emph{Similar}.

\subsection{SAP model}

While SAP \cite{bib:castro2022unconstrained} is convex by construction, it is a
good exercise to verify that our new conditions hold for this model. SAP
regularizes friction according to
\begin{eqnarray}
    \bgamma_t(\vf{v}_c) &=& -\min\left(\frac{\Vert\vf{v}_t\Vert}{R_t}, \mu\gamma_n(v_n)\right)\hat{\vf{t}}
    \label{eq:sap_friction_regularization}
\end{eqnarray}
where $R_t$ is SAP's regularization
parameter~\cite{bib:castro2022unconstrained}. From
\cite{bib:castro2022unconstrained}, we know that the Hessian of the regularizer
cost
\begin{equation*}
    \mf{G} = \frac{\partial^2\ell}{\partial\mf{v}_c^2} = -\frac{\partial\bgamma}{\partial\mf{v}_c}=
    -\begin{bmatrix}
		\frac{\partial\bgamma_t}{\partial\mf{v}_t} & \frac{\partial\bgamma_t}{\partial v_n}\\
		\frac{\partial\gamma_n}{\partial\mf{v}_t}^T & \frac{\partial\gamma_n}{\partial v_n}\\
	\end{bmatrix}
\end{equation*}
is symmetric positive semi-definite and therefore SAP satisfies condition
\eqref{eq:curl_tangent}. Moreover, $\partial\bgamma_t/\partial\mf{v}_t$ is
symmetric since $\bgamma_t$ is of the isotropic form
\eqref{eq:generic_friction_model} and condition \eqref{eq:curl_normal} is met
also.

\subsection{Lagged Model}
\label{sec:lagged_model}

We use the model of regularized friction \eqref{eq:regularized_friction} in
which the normal impulse is \emph{lagged} to the previous time step
\begin{equation}
    \bgamma_t(\vf{v}_t)=-\mu\,f(s)\gamma_{n,0}\hat{\mf{t}}
    \label{eq:lagged_friction_model}
\end{equation}
with $s=\Vert\mf{v}_t\Vert/\varepsilon_s$, and using
\eqref{eq:generic_force_law}, $\gamma_{n,0}=\delta t f_n(\phi_0, v_{n,0})$. For a
physical model of compliance for which $\gamma_n$ is only a function of $v_n$,
condition \eqref{eq:curl_tangent} is trivially met since
\begin{eqnarray*}
    \frac{\partial\bgamma_t}{\partial v_n}=\mf{0}\\
    \frac{\partial\gamma_n}{\partial\mf{v}_t}=\mf{0}
\end{eqnarray*}
which implies that the contact potential is separable as the sum of normal and
friction contributions
\begin{equation*}
    \ell(\vf{v}_c) = \ell_t(\vf{v}_t) + \ell_n(v_n).
\end{equation*}

Notice that even though they are lagged in the friction component, the normal
impulses $\gamma_n(v_n)$ are still treated implicitly, with normal potential
$\ell_n(v_n)$.

We verify that
\begin{equation*}
    \ell_t(\vf{v}_t) = \mu\gamma_{n0}\,\varepsilon_s\,F(\Vert\vf{v}_t\Vert/\varepsilon_s)
\end{equation*}
with $f = F'$ satisfies $\bgamma_t = -\partial\ell_t/\partial\mf{v}_t$. 

The Hessian of $\ell_t$ is given by
\begin{eqnarray*}
    \frac{\partial^2\ell_t}{\partial\vf{v}_t^2}&=&-\frac{\partial\bgamma_t}{\partial\vf{v}_t}\nonumber\\
    &=&\mu\gamma_{n0}\left[\frac{f'(s)}{\varepsilon_s}\mf{P}(\hat{\vf{t}})+\frac{f(s)}{\Vert\vf{v}_t\Vert}\mf{P}^\perp(\hat{\vf{t}})\right].
\end{eqnarray*}

With a non-decreasing $f$, $F$ is convex, and the Hessian of $\ell_t$ is the
positive linear combination of two projection matrices. Therefore, $\ell_t$ is
twice differentiable and convex. Thus the Hessian
\begin{equation*}
    \frac{\partial^2\ell}{\partial\mf{v}_c^2} = -\frac{\partial\bgamma}{\partial\mf{v}_c}=
    \begin{bmatrix}
		-\frac{\partial\bgamma_t}{\partial\mf{v}_t} & \vf{0}\\
		\vf{0}^T & -\frac{\partial\gamma_n}{\partial v_n}\\
	\end{bmatrix}
\end{equation*}
is positive definite for physics-based models that satisfy
\eqref{eq:normal_sufficient_conditions}.

With the judicious choice $F(s)=\sqrt{s^2+1}-1$ from Section
\ref{sec:compliant_contact}, the cost, gradient, and Hessian involve soft norms
(Appendix \ref{sec:soft_norms}) which are twice differentiable and convex, even
at $\vf{v}_t=\vf{0}$
\begin{eqnarray*}
    \ell_t(\vf{v}_t) &= \mu\gamma_{n0}\,\Vert\vf{v}_t\Vert_s,\nonumber\\
    \bgamma_t&=-\mu\,\gamma_{n,0}\hat{\mf{t}}_s,\nonumber\\
    \frac{\partial^2\ell_t}{\partial\vf{v}_t^2}&=\mu\gamma_{n0}\frac{\mf{P}^\perp(\hat{\vf{t}}_{s})}{\Vert\vf{v}_t\Vert_s+\varepsilon_s}.
\end{eqnarray*}

\reviewquestion{Q5}{A remarkable property of this approximation is that it
completely eliminates the \emph{gliding} artifact characteristic of previous
convex formulations. This is discussed in detail in Section
\ref{sec:gliding_artifact}.}

\subsection{Similar Model}
\label{sec:similar_model}

Similarity solutions to partial differential equations (PDEs) are solutions that depend on certain groupings of
independent variables rather than each variable individually. In particular,
self-similar solutions arise when the problem lacks a characteristic time or
length scale. The Blasius solution to Prandtl's boundary layer equations in
fluid mechanics is a well-known and celebrated example.

 Motivated by the algebraic form of SAP impulses
\cite{bib:castro2022unconstrained}, we propose the grouping of variables $z =
v_n-\mu\Vert\vf{v}_t\Vert$. Furthermore, we generalize this grouping to
\begin{equation}
    z = v_n-\mu \varepsilon_sF(s),
    \label{eq:similar_grouping}
\end{equation}
where, as in Section \ref{sec:lagged_model}, $\mu \varepsilon_sF(s)$ simplifies
to $\mu\Vert\vf{v}_t\Vert_s$ when $F(s)=\sqrt{s^2+1}-1$. Note the consistency of
units in \eqref{eq:similar_grouping}, an important aspect of similar solutions.
With this grouping, we propose the \emph{similar} solution
\begin{eqnarray}
    \gamma_n(\Vert\vf{v}_t\Vert,v_n) &=& n(z),\nonumber\\
    \bgamma_t(\Vert\vf{v}_t\Vert,v_n) &=&
    -\mu\,f(s)\,\gamma_n(\Vert\vf{v}_t\Vert,v_n)\,\hat{\vf{t}}.
    \label{eq:impulses_convexified}
\end{eqnarray}

Unlike the Lagged model, the Similar model strongly couples friction and normal
components. However, this introduces a dependency of the normal component on
slip speed, an artifact we quantify in the following sections.

Differentiation of \eqref{eq:impulses_convexified} leads to
\begin{eqnarray*}
    \frac{\partial\bgamma_t}{\partial
    v_n}=\frac{\partial\gamma_n}{\partial\mf{v}_t}=-\mu\,f(s)\,n'(z)\,\hat{\vf{t}},
\end{eqnarray*}
which confirms condition \eqref{eq:curl_tangent}. To find the potential, we
start from the normal component of the impulse
\begin{equation*}
    \gamma_n(\Vert\vf{v}_t\Vert,v_n) = n(z) = -\frac{\partial\ell}{\partial v_n},
\end{equation*}
and integrate on $v_n$ to obtain
\begin{equation*}
    \ell(\vf{v}_t,v_n) = -N(z) + G(\vf{v}_t),
\end{equation*}
where $G(\vf{v}_t)$ is an arbitrary function of $\vf{v}_t$. Taking the
derivative with respect to $\vf{v}_t$ results in
\begin{equation*}
    \frac{\partial\ell}{\partial\vf{v}_t}=\mu\,n(z)\,f(s)\hat{\vf{t}} + \frac{\partial G}{\partial\vf{v}_t}.
\end{equation*}

Comparing this equation with \eqref{eq:impulses_convexified} reveals that we can
set $G=0$ and obtain
\begin{equation}
    \ell(\Vert\vf{v}_t\Vert,v_n) = -N(z),
    \label{eq:similar_potential}
\end{equation}
as the desired potential function.

The Hessian of this potential is
\begin{eqnarray*}
    &&\frac{\partial^2\ell}{\partial\vf{v}_c^2}=-\frac{\partial\bgamma}{\partial\vf{v}_c}=\\
    &&\begin{split} =&\mu\left[
        \left(\frac{f'(s)n(z)}{\varepsilon_s}-n'(z)f^2(s)\right)\mf{P}(\hat{\vf{t}})\right.\\
    &+\left.\frac{f(s)n(z)}{\Vert\vf{v}_t\Vert}\mf{P}^\perp(\hat{\vf{t}})\right].
    \end{split}\nonumber
\end{eqnarray*}

For a convex potential $\ell_n$ we have $n'=N''=-\ell_n'' \leq 0$. With $f(s)$
non-decreasing as in the lagged model, $f'\ge 0$. The Hessian
$\partial^2\ell/\partial\vf{v}_c^2$ is the linear combination with positive
coefficients of symmetric positive semi-definite projection matrices. Therefore,
the Hessian is symmetric positive semi-definite, and the potential is convex.
\reviewquestion{Q29}{As with Lagged,} the choice $F(s)=\sqrt{s^2+1}-1$ leads to
continuously differentiable expressions in terms of soft norms (Appendix
\ref{sec:soft_norms}), with no singularity at $\vf{v}_t=\vf{0}$.

\begin{figure}[!h]
    \centering
    \includegraphics[width=0.6\columnwidth]{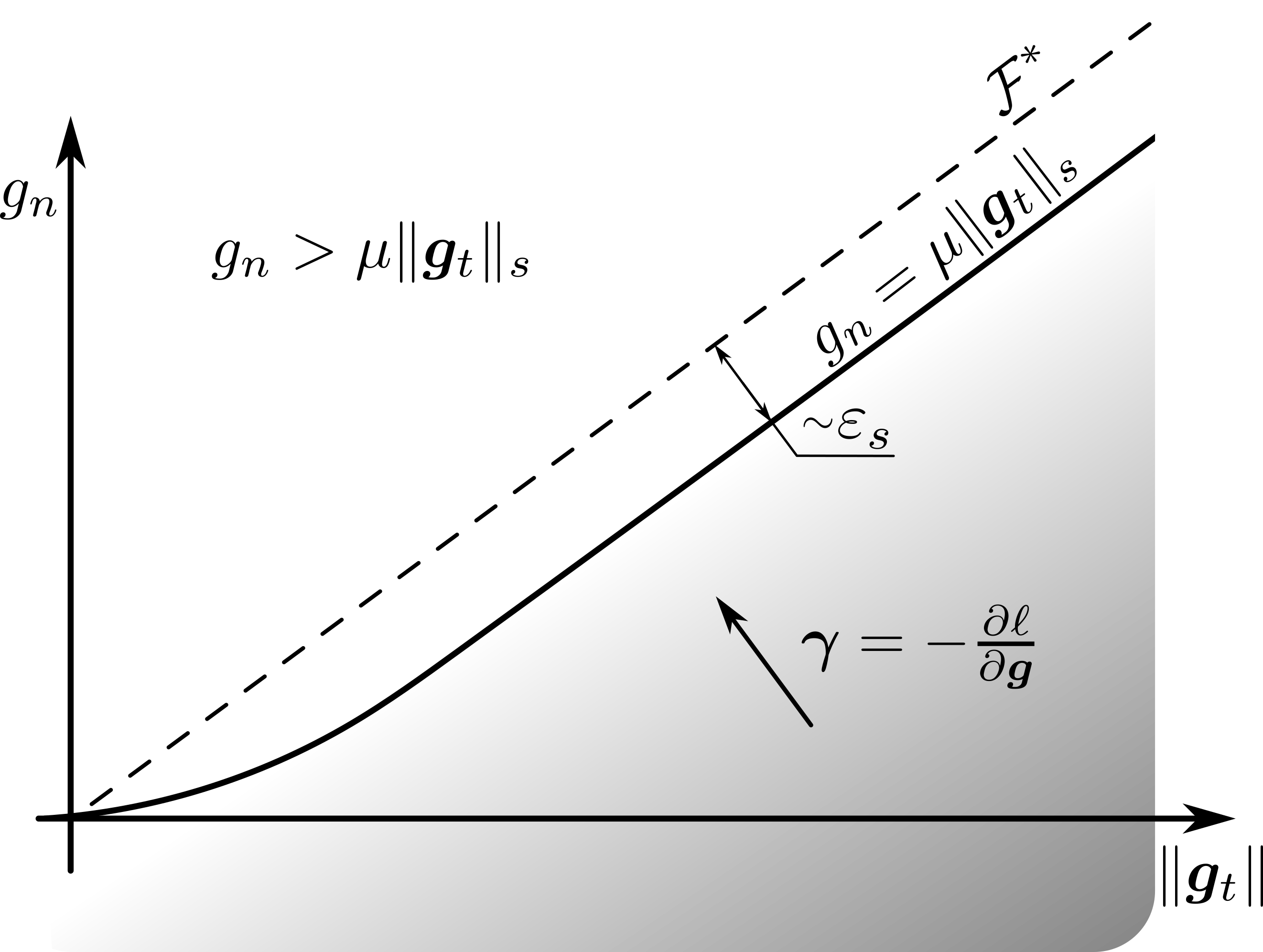}
    \caption{The \emph{Similar} model penalizes $\vf{g}$ when it lies outside
    the region $g_n \ge \mu \varepsilon_sF(\Vert\vf{g}_t\Vert/\varepsilon_s)$.
    With $F(s)=\sqrt{s^2+1}-1$ and $\varepsilon_s\rightarrow 0$, the model
    enforces $\vf{g} \in \mathcal{F}^*$ in the limit to rigid contact,
    consistent with the primal formulations in
    \cite{bib:mazhar2014,bib:castro2022unconstrained}.}
    \label{fig:dual_cone}
\end{figure}

Our \emph{Similar} model closely relates to the primal formulation
\cite{bib:mazhar2014,bib:castro2022unconstrained} of Anitescu's convex
approximation \cite{bib:anitescu2006}. Following
\cite{bib:castro2022unconstrained} we define the velocity $\vf{g} = \vf{v}_c -
\hat{\vf{v}}_c$, with $\hat{\vf{v}}_c=[0, 0, \hat{v}]^T$ and the \emph{breaking
velocity} from \eqref{eq:breaking_velocity}. Using this definition, we express $z$
as $z = g_n-\mu \varepsilon_sF(\Vert\vf{g}_t\Vert/\varepsilon_s) + \hat{v}$. We
notice that $\ell(z)$ is constant (no contact) for $z\ge\hat{v}$. In terms of
$\vf{g}$, this is equivalent to $g_n-\mu
\varepsilon_sF(\Vert\vf{g}_t\Vert/\varepsilon_s) \ge 0$. The graph of $g_n=\mu
\varepsilon_sF(\Vert\vf{g}_t\Vert/\varepsilon_s)$ defines the boundary of the
stick-slip transition, and therefore when $F(s)$ is convex, its epigraph (the
set of points above its graph), corresponds to a convex region (not necessarily
a cone as in previous work). This is illustrated in Fig. \ref{fig:dual_cone}. In
this plot, contour levels of $\ell(\Vert\vf{g}_t\Vert,g_n)$ correspond to lines
of constant $z$, which by definition are perpendicular to the gradient
$\partial\ell/\partial\mf{v}_c$. We see that the role of the potential $\ell$ is
to penalize $\vf{g}$ when it lies outside the epigraph of $g_n=\mu
\varepsilon_sF(\Vert\vf{g}_t\Vert/\varepsilon_s)$.

Consider $F(s)=\sqrt{s^2+1}-1$, for which $g_n = \mu \Vert\vf{g}_t\Vert_s$. The
epigraph of this function is an approximation to the dual $\mathcal{F}^*$ of the
friction cone $\mathcal{F}$, Fig. \ref{fig:dual_cone}. This approximation is
$\mathcal{F}^*$ in the limit $\varepsilon_s\rightarrow 0$. Moreover, in the
limit to rigid contact, $\ell$ enforces $\vf{g} \in \mathcal{F}^*$, which
corresponds to the cone constraint in the primal formulation
\cite{bib:mazhar2014,bib:castro2022unconstrained}.

\section{Convex Approximation of Compliant contact with Hunt \& Crossley Dissipation}
\label{sec:convex_hunt_crossley}

So far, we have considered a generic functional form of the compliant law in
\eqref{eq:generic_force_law}. To close our model, we consider a linear elastic
law in the signed distance $\phi$ with Hunt \& Crossley~\cite{bib:hunt_crossley}
dissipation. As long as conditions \eqref{eq:normal_sufficient_conditions} are
met, other alternatives such as the Hertz model can be used. In practice,
however, users often find the linear model satisfactory. We write this model
within framework~\eqref{eq:generic_force_law} as
\begin{equation}
    f_n(\phi, v_n) = k\,(-\phi)_+\,(1-dv_n)_+,
    \label{eq:linear_compliance_with_hunt_crossley}
\end{equation}
where $k$ is the linear contact stiffness, $d$ is the Hunt \& Crossley
dissipation parameter and $(a)_+ = \max(0, a)$. This force is zero whenever $v_n
\ge \hat{v}$, with
\begin{equation}
    \hat{v}=\min(-\phi_0/\delta t, 1/d),
    \label{eq:breaking_velocity}
\end{equation}
the minimum normal velocity for contact to break. For $v_n < \hat{v}$ we define
the (indefinite) antiderivative $N^+(v_n)$
\begin{align*}
    N^+(v_n; \phi_0) &= \\
    \delta t\,k\,&\left[-v_n\left(\phi_0 + \frac{1}{2}\delta t\,v_n\right) + d\,\frac{v_n^2}{2}\left(\phi_0 + \frac{2}{3}\delta t\,v_n\right)\right].
\end{align*}


Since the impulse is zero for $v_n \ge \hat{v}$, its antiderivative must be
constant. Therefore, we write
\begin{equation*}
    N(v_n) = N^+(\min(v_n, \hat{v}); f_{0}),
\end{equation*}
resulting in a continuous function for all values of $v_n$.

Though inherently compliant, this model effectively handles very stiff contact
due to the robustness of our convex formulation. In Section~\ref{sec:results},
we use Hertz theory to estimate steel stiffness, and
Section~\ref{sec:applications} demonstrates that our approach can manage
stiffnesses up to five orders of magnitude higher.

Finally, we observe that this model is easily extended to incorporate
\emph{hydroelastic contact}~\cite{bib:elandt2019pressure} --- a modern rendition
of \emph{Elastic Foundation}~\cite[\S 4.3]{bib:johnson1987_contact_mechanics} to
model continuous contact patches. We use the approach
in~\cite{bib:masterjohn2021discrete}.
\section{Comparative Analysis of the Convex Approximations}
\label{sec:comparative_analysis}

All models in Section \ref{sec:models_family} are convex approximations of
contact, each with its own strengths and limitations. Hereinafter, we refer to
each of these formulations as SAP \cite{bib:castro2022unconstrained},
\emph{Lagged} (section \ref{sec:lagged_model}), and \emph{Similar} (section
\ref{sec:similar_model}). We summarize their properties and artifacts in Table
\ref{tab:models_properties} and analyze them in detail in the following
subsections. For the Lagged and Similar models, we use the compliant model with
Hunt \& Crossley dissipation from Section \ref{sec:convex_hunt_crossley}.

\begin{table*}[h]
    \centering
    \begin{tabular}{c|c|c|c|cc|c}
                                     Model & Consistent  & Strongly coupled &
                                     Gliding during slip   & \multicolumn{2}{|c|}{Compliance
                                     modulation} & Transition slip \\
                                     & & & & Stiffness & Dissipation \\
        \hline
        SAP           & No  & Yes & $\mu(\delta t + \tau_d)\Vert\vf{v}_t\Vert$ &
        $k/(1+\tilde{\mu}^2)$ &$\tau_d$ & $\sigma\,\text{w}\,\mu\,\gamma_n$\\
        Lagged        & Yes & No  &  ---                                       &
        --- & --- & $v_s$ \\
        Similar       & No  & Yes & $\mu\delta t\Vert\vf{v}_t\Vert$            &
        $k\,(1+\mu\,d\,\Vert\vf{v}_t\Vert)$ & $d/(1+\mu\,d\,\Vert\vf{v}_t\Vert)$
        & $v_s$ \\
    \end{tabular}
    \caption{Properties of each convex approximation in Section \ref{sec:models_family}.}
    \label{tab:models_properties}
\end{table*}

\subsection{Gliding During Slip}
\label{sec:gliding_artifact}

SAP, as well as the formulations from Anitescu and Todorov
\cite{bib:anitescu2006,bib:todorov2011}, introduce a number of artifacts as the result
of the convex approximation of contact. The most well-known artifact is that
objects \emph{glide} at a finite offset during slip \cite{bib:mazhar2014,
bib:horak2019, bib:castro2022unconstrained}. For SAP,
this offset is \cite{bib:castro2022unconstrained}
\begin{equation}
    \phi_\text{offset} = \mu(\delta t + \tau_d)\Vert\vf{v}_t\Vert.
    \label{eq:sap_gliding_offset}
\end{equation}
With zero dissipation ($\tau_d=0$), this reduces to the gliding distance $\mu\delta
t\Vert\vf{v}_t\Vert$ in Anitescu's formulation of rigid contact
\cite{bib:anitescu2006}. There are no artifacts during stiction.

While the term $\mu\delta t\Vert\vf{v}_t\Vert$ can be negligible for small time
steps and slip speeds, the term $\mu\tau_d\Vert\vf{v}_t\Vert$ can dominate when
dissipation is large. Therefore, users must face the trade-off between accurate
dissipation modeling and the magnitude of this gliding artifact, making the
choice of this parameter cumbersome.

\reviewquestion{Q5}{The Similar approximation eliminates this coupling with
dissipation, though it still introduces the gliding artifact. This is caused by the
non-physical coupling of the tangential velocity into the normal component of the
contact impulse in \eqref{eq:impulses_convexified}. Replacing $v_n$ with $z$
from \eqref{eq:similar_grouping} into
\eqref{eq:linear_compliance_with_hunt_crossley}, we find that the impulse goes
to zero at $\phi_\text{offset} = \mu\delta t\Vert\vf{v}_t\Vert$ rather than at
$\phi=0$. Gliding during slip for SAP can be traced back to this same artificial
coupling of the tangential velocity into the normal force; see
\cite{bib:castro2022unconstrained} for details.

In contrast, the Lagged approximation does not introduce gliding, since it
completely eliminates the contribution of the tangential velocity in the normal
force component.}

\subsection{Compliance Modulation}
\label{sec:compliance_modulation}

For SAP, the effective compliance during slip is $k_\text{eff} =
k/(1+\tilde{\mu}^2)$, where $\tilde{\mu}=\mu(R_t/R_n)^{1/2}$
\cite{bib:castro2022unconstrained} depends on the ratio of the tangential to
normal regularization parameters. SAP chooses this ratio so that
$\tilde{\mu}\approx0$, but it is still non-zero. We refer to this artifact as
\emph{compliance modulation}, and it was first reported in
\cite{bib:castro2022unconstrained}.

Replacing $v_n$ with $z$ in \eqref{eq:linear_compliance_with_hunt_crossley} and
factoring out the term $1+\mu\,d\,\Vert\vf{v}_t\Vert$, we see that during slip,
the Similar approximation has effective stiffness $k_\text{eff} =
k\,(1+\mu\,d\,\Vert\vf{v}_t\Vert)$ and dissipation $d_\text{eff} =
d/(1+\mu\,d\,\Vert\vf{v}_t\Vert)$. That is, the effective stiffness increases
while the effective dissipation decreases during slip.

The Lagged approximation does not introduce compliance modulation.

\subsection{Strong Coupling}
\label{sec:strong_coupling}

\reviewquestion{Q18}{We use the term \emph{strong coupling} to describe the
tight, algebraic, coupling of the friction and normal components. While SAP and
Similar approximations do introduce strong coupling, the Lagged approximation
lags the normal force in Coulomb's law.} Our numerical studies (Sections
\ref{sec:results} and \ref{sec:applications}) show that this lag effect is
noticeable only in high-energy impacts where normal forces change rapidly.
Otherwise, the approximation is $\mathcal{O}(\delta t)$, consistent with the
first-order Taylor expansion of the signed distance introduced in Section
\ref{sec:compliant_contact}. Moreover, our grasp stability analysis (Section
\ref{sec:franka_hand}) confirms that the lagged approximation is suitable, even
for highly dynamic manipulation tasks.

\subsection{Consistency}
\label{sec:consistency}

In the context of ordinary differential equations (ODEs), a scheme is
\emph{consistent} when the original set of ODEs is recovered in the limit $\delta
t\rightarrow 0$. This is not the case for SAP and Similar models, due to the
existence of \emph{compliance modulation} (Section
\ref{sec:compliance_modulation}), regardless of the time step size. In particular
for SAP, the gliding artifact does not vanish in \eqref{eq:sap_gliding_offset}
unless dissipation is zero. Conversely, Lagged is a consistent first order
approximation of \eqref{eq:total_impulse}.

\subsection{Stick-Slip Transition}
\label{sec:stiction_tolerance}

Since the models in this work \emph{regularize} friction, the transition
between stiction and sliding occurs at a finite slip velocity. For Lagged and
Similar models, this transition is parameterized by $\varepsilon_s$ in
\eqref{eq:regularized_friction}, and it is fixed by the user to a value $v_s$
which we call the \emph{stiction tolerance}.

For SAP, we see from \eqref{eq:sap_friction_regularization} that stick-slip
transitions occur at a magnitude of slip equal to
\begin{equation}
    v_s = \sigma\,\text{w}\,\mu\,\gamma_n,
    \label{eq:sap_stiction_tolerance}
\end{equation}
where we used the fact that SAP uses $R_t = \sigma\,\text{w}$, with $\text{w}$ a
diagonal approximation of the Delassus operator
\cite{bib:castro2022unconstrained} (the \emph{effective inverse mass} of the
contact). The dimensionless parameter $\sigma$ is set to $\sigma=10^{-3}$, a
good trade-off between numerical conditioning and a tight approximation of
stiction, as required for the simulation of manipulation applications in
robotics. Refer to \cite{bib:castro2022unconstrained} for a thorough study of
SAP's properties.

\section{Impacts, Stiffness and Conditioning}
\label{sec:impacts_and_conditioning}

We are interested in the numerical \emph{stiffness} introduced by the
regularization of friction as it is critical for a robust numerical
implementation. We define $G_t= d\Vert\bgamma_t\Vert/d\Vert\vf{v}_t\Vert$ and
examine its value at $\Vert\vf{v}_t\Vert=0$. For all approximations we find
\begin{equation*}
    G_t = \frac{\mu}{v_s}\tilde{\gamma}_n,
\end{equation*}
with $v_s$ the stiction tolerance (Table \ref{tab:models_properties}),
$\tilde{\gamma}_n =\gamma_n$ for SAP and Similar models, and $\tilde{\gamma}_n
=\gamma_{n,0}$ for Lagged.

As seen in \eqref{eq:sap_stiction_tolerance}, SAP's stiction tolerance depends
on impulse, often much higher during impacts than in sustained contact
(e.g., a robot grasping an object). This effectively increases SAP's
regularization during impacts, making the problem better-conditioned but
reducing stiction accuracy. In contrast, Lagged and Similar solve a much stiffer
problem given the stiction tolerance $v_s$ is constant. We confirm this in
Section \ref{sec:clutter} in a case with impacts.

The situation reverses during sustained contact, as the normal impulse depends
mainly on object's weight, and external forces, not rapid velocity changes.
Since impulse scales with time step size, $G_t = \mathcal{O}(\delta t)$ for
Similar and Lagged, while SAP's $G_t = 1/R_t$ remains constant (with $R_t$ from
\eqref{eq:sap_friction_regularization}), leading to a stiffer problem. This is
confirmed in Section \ref{sec:clutter} as we analyze the numerical conditioning
of the problem and the solver's number of iterations.

It could be argued that, for most robotics applications,
accurately resolving stiction during impacts is unnecessary. Therefore, we
propose a \emph{regularized} version of the Lagged model, where the
regularization parameter is
\begin{equation}
    \varepsilon_s = \max(v_s, \sigma\,\text{w}\,\mu\,\gamma_{n, 0}).
    \label{eq:regularized_vs}
\end{equation}
In this formulation, regularization is \emph{softened} during strong impacts when
values of $\gamma_{n, 0}$ are higher, but a tight value bounded by $v_s$ is used
for sustained contacts. We study the effect of this formulation in Section
\ref{sec:clutter}.

\section{Barrier Functions As Compliant Contact}
\label{sec:barriers_and_compliance}

\reviewquestion{Q3}{While in this work we find that compliance provides a
reliable approximation of contact, we cannot neglect the fact that most
simulation software in the robotics and graphics communities implement rigid
approximations of contact. It is important to point out that rigid contact, as
well as compliant contact, are both approximations of the real physics, each with
their strengths and limitations.

In this section, we provide a discussion on the connection between these two
apparently unrelated technologies, in particular when barrier or similarly
smooth functions are used to enforce the non-penetration and complementarity
conditions appearing in rigid approximations \cite{bib:howell2022,
bib:li2020ipc, bib:macklin2019}. We have not implemented any of these barriers
in our framework, though this could be an interesting avenue of future
research.}

\subsection{Logarithmic Barriers for Interior Point Methods}

Interior-point methods (IP) are the standard for solving optimization problems
with inequality constraints, essential for rigid contact modeling. IP solvers
can also handle a richer set of constraints, like conic constraints, as needed
to model Coulomb friction. Notable solvers include open-source
Ipopt~\cite{bib:wachter2006} and proprietary options like
Gurobi~\cite{bib:gurobi} and Mosek~\cite{bib:mosek}. Fundamentally, IP methods
replace constraints with a barrier function that penalizes infeasible solutions.
Logarithmic functions are commonly used, which for rigid contact penalize
distances near zero
\begin{equation*}
    \ell_n(\phi) = -\kappa \ln(\phi),
\end{equation*}
while penetration states ($\phi\le0$) are infeasible. The barrier parameter
$\kappa > 0$ is iteratively reduced to approach a rigid approximation, though it
can never reach zero. Therefore, in practice the solution is effectively
compliant, with a force law that fits the framework
\eqref{eq:generic_force_law}
\begin{equation}
    f_n(\phi, v_n) = \frac{\kappa}{\phi}\quad  \phi \ge 0.
    \label{eq:logarithmic_barrier_force}
\end{equation}

As a physical model of contact, users may struggle to
interpret~\eqref{eq:logarithmic_barrier_force}, which has a parameter $\kappa$
with the unit of energy and nonphysical action at infinite distance. In
practice, $\kappa$ is not exposed, but hidden as part of the solver internals.
Users \emph{believe} they are working with a true model of rigid contact when,
in reality, the solver is using an compliant model approximation. Even if
$\kappa$ can be reduced to very small values (according to some hidden metric),
the solver often ends up solving a much more challenging problem than needed
since, in reality, physical materials have finite stiffness.

\subsection{Incremental Potential Contact}

IPC ~\cite{bib:li2020ipc} is an optimization-based framework to model rigid
contact and guarantee intersection-free solutions. The method attains strong
robustness given its implicit time-stepping scheme and line
search augmented with continuous collision detection (CCD) to maintain
feasibility. However, IPC formulates a non-convex optimization problem and can
fall into local minima, not satisfying the original physical laws. Moreover, the
method lacks convergence guarantees, as properly pointed out by the original
authors.

Similar to the logarithmic barrier functions of IP methods, IPC proposes a $C^2$
potential
\begin{eqnarray*}
    \ell_c(\phi) = -\kappa_\text{IPC}\,(\hat{d}-\phi)_+^2\ln(\phi/\hat{d})
\end{eqnarray*}
where $\kappa_\text{IPC}$ is a parameter automatically adjusted to improve
numerical conditioning, and $\hat{d}$ is a user parameter. Typical values of
$\hat{d}$ used by the original authors in their extensive simulations test cases
are in the range $0.1\text{ mm}$ to $1\text{ mm}$. This potential is proposed to
achieve intersection-free solutions, eliminate nonphysical action at infinite
distances, and maintain smoothness for better numerics. We observe that this
method again fits the framework \eqref{eq:generic_force_law}, with a
\emph{compliant} contact force law of the form
\begin{equation*}
    f_n(\phi, v_n) = \kappa_\text{IPC}\,(\hat{d}-\phi)_+\left[\frac{(\hat{d}-\phi)_+}{\phi} - 2\ln(\phi/\hat{d})\right],
\end{equation*}
which is only non-zero for $\phi\in(0, \hat{d}]$. Performing Taylor expansion around
$\phi=\hat{d}$, we see that $f_n\approx
3\kappa_\text{IPC}(\hat{d}-\phi)_+^2/\hat{d}$, and the force models a quadratic spring of
stiffness $\kappa_\text{IPC}$ (with units of N/m). In the limit to
$\phi\rightarrow 0^+$, the force approximates
$f_n\approx\kappa_\text{IPC}\hat{d}^2/\phi$, the interior point force
\eqref{eq:logarithmic_barrier_force}.

In summary, this method models a thin, compliant layer around a rigid core
instead of the strict non-penetration conditions largely favored in the
literature. We do not view this as a flaw, as the authors have demonstrated their method's
effectiveness through extensive simulation studies. Instead, we see this as
an indication of the levels of rigidity that can be achieved in practice.

\section{Implementation}
\label{sec:implementation}

Our work is implemented in Drake \cite{bib:drake}, a robotics toolkit that
provides modeling abstractions and optimization tools for the modeling,
simulation, and analysis of robotics systems. Our implementation includes support
for deformable Finite Element Models (FEM), holonomic constraints, PD
controllers with effort limits, reflected inertia, and the modeling of
continuous contact patches with \emph{Hydroelastic Contact}. Refer to our
previous work for further details
\cite{bib:castro2022unconstrained,bib:masterjohn2021discrete,bib:han2023}.

SAP \cite{bib:castro2022unconstrained} uses Newton's method to compute a search
direction $\Delta\mf{v}$ for \eqref{eq:unconstrained_fomulation} according to
\begin{equation}
    \mf{H}(\mf{v})\Delta\mf{v} = -\mf{r}(\mf{v}),
    \label{eq:search_direction}
\end{equation}
where $\mf{r}(\mf{v})=\partial\ell/\partial\mf{v}$ is the \emph{residual} of the
primal cost $\ell(\mf{v})$, and $\mf{H}(\mf{v})$ is its Hessian. Upon convergence,
the residual $\mf{r}(\mf{v})$ is zero, which is equivalent to the linearized
balance of momentum \eqref{eq:optimality_condition}. The solution at iteration
$m$ is updated as
\begin{equation*}
    \mf{v}_{m+1} = \mf{v}_{m} + \alpha\Delta\mf{v},
\end{equation*}
with $\alpha$ determined via a line search along $\Delta\mf{v}$ to minimize
$\ell(\mf{v})$. Our line search uses a Newton-Raphson method based on \cite[\S
9.4]{bib:numerical_recipes}, augmented with bracketing and bisection to
guarantee convergence. The derivatives required for the line search are computed with
$\mathcal{O}(n)$ complexity \cite{bib:castro2022unconstrained}, enabling
convergence to machine precision with negligible computational overhead.
This strategy has proven to be very robust in practice.

Multibody \emph{tree} structures create distinct cliques, while degrees of
freedom (DOFs) for modeling a deformable body are also grouped into their own
cliques \cite{bib:han2023}. Similarly, constraints involving the same pair of
cliques are grouped into \emph{clusters}. We exploit this structure using a
supernodal Cholesky factorization \cite[\S 9]{bib:davis2016survey} of the
Hessian $\mf{H}$ in \eqref{eq:search_direction} that takes advantage of dense
algebra optimizations. We compute the elimination ordering of the supernodes using
approximate minimum degree (AMD) ordering \cite{bib:amestoy1996approximate} to
minimize fill-ins.

\subsection{Geometric Queries}
\label{sec:geometry}

\reviewquestion{Q6}{Unlike other approaches, our method performs only a single
geometric query at the start of each time step, avoiding repeated queries during
Newton iterations. Changes in contact configuration are captured using the
first-order approximation of the signed distance discussed in
Section~\ref{sec:compliant_contact}. This approximation is extended in
\cite{bib:masterjohn2021discrete} to model continuous contact surfaces with
hydroelastic contact \cite{bib:elandt2019pressure}, implemented as part of Drake
and accelerated with an oriented bounding box hierarchy. Point contact queries
use FCL \cite{bib:fcl}. All simulations in this work include contact pairs with
signed distance below a fixed threshold, $\phi < 10$~cm, including contact pairs
even before physical contact is established.}

\subsection{Differentiation Through Contact}
\label{sec:gradients}

Our implementation in Drake provides an end-to-end solution for the computation
of gradients through contact for applications such as system identification,
reinforcement learning, and trajectory optimization. \reviewquestion{Q13}{While
differentiation through contact is not new \cite{bib:geilinger2020,
bib:howell2022, bib:lelidec2025}, our approach is unique in that we use
automatic differentiation to compute gradients through geometry queries and the
dynamics ($\partial\mf{r}/\partial\btheta$ below) and the \emph{implicit
function theorem} to propagate these gradients to the next time solution.}

We denote differentiation parameters with $\btheta\in\mathbb{R}^{n_\theta}$,
which can include physical quantities such as mass and inertias, contact
parameters, actuation, and even the previous state of the multibody system.
Using this notation, SAP's optimality condition can be written in terms of the
residual in \eqref{eq:search_direction} as $\mf{r}(\mf{v}; \btheta) = \mf{0}$.
In this notation, the residual is a function of the
generalized velocities $\mf{v}$ \emph{given} a set of parameters $\btheta$.

We use the implicit function theorem on $\mf{r}(\mf{v}; \btheta) = \mf{0}$ 
\begin{equation*}
    \frac{\partial \mf{r}(\mf{v}; \btheta)}{\partial\mf{v}}\frac{d\mf{v}}{d\btheta} + \frac{\partial \mf{r}(\mf{v}; \btheta)}{\partial\btheta} = \mf{0},
\end{equation*}
and note that $\partial\mf{r}/\partial\mf{v}=\mf{H}(\mf{v})$ to write
\begin{equation}
    \mf{H}\frac{d\mf{v}}{d\btheta} = -\frac{\partial \mf{r}(\mf{v}; \btheta)}{\partial\btheta}.
    \label{eq:v_derivatives}
\end{equation}

In our approach, the expensive-to-compute Cholesky factorization of $\mf{H}$ is
only computed at each Newton iteration in \eqref{eq:search_direction} during
forward simulation. Upon convergence, $\mf{H}$ is already assembled and
factorized and is thus reused in \eqref{eq:v_derivatives} an additional
$n_\theta$ times to propagate derivatives through the solver into gradients
$d\mf{v}/d\btheta$ of the generalized velocities.

In our hybrid approach, $\partial\mf{r}/\partial\btheta$ in
\eqref{eq:v_derivatives} is computed with automatic differentiation. This
enables the computation of gradients through arbitrarily complex geometric
models, while the implicit function theorem propagates derivatives accurately
and efficiently through the contact resolution phase.

\section{Test Cases}
\label{sec:results}

We analyze a series of two-dimensional cases to assess accuracy, quantify
artifacts introduced by the convex approximations, and gain intuition into the
physics and numerics. \reviewquestion{Q31}{Here, we focus on comparing our
convex formulations. We refer the reader to \cite{bib:horak2019,bib:lelidec2024}
for previous work comparing other approaches, including Anitescu's convex
formulation.}

For all cases, we use $v_s=10^{-4}\text{ m}/\text{s}$ for Lagged and Similar,
and $\sigma=10^{-3}$ for SAP, leading to very tight stiction modeling as
required for simulating manipulation tasks.

We estimate contact stiffness using Hertz theory. For a sphere of mass $m$ and
radius $R$, Hertz theory predicts a penetration
$\delta=(3mg/(4ER^{1/2}))^{2/3}$. For steel with Young's modulus $E=200\text{
GPa}$ and using the radii and masses from Sections \ref{sec:falling_sphere} and
\ref{sec:sliding_rod}, we obtain penetrations around $\delta\approx
2.5\times10^{-7}\text{ m}$ and stiffnesses $k\approx 1\times
10^{7}-2\times 10^{7}\text{ N}/\text{m}$. We use $k = 10^{7}\text{
N}/\text{m}$ for all cases in this section.

For some cases, we perform a convergence study where we compute the
error in the positions $\mf{q}_{\delta t}$ obtained using step size $\delta t$
against a reference $\mf{q}_\text{ref}$ as a function of the time step size
\begin{equation*}
    e_q(\delta t) = \left(\frac{1}{T}\int_0^T dt\Vert\mf{q}_{\delta t}(t)-\mf{q}_\text{ref}(t)\Vert^2\right)^{1/2}
\end{equation*}
where $T$ is simulation duration. The reference solution is obtained numerically
using a time step 10 times smaller than the smallest time step in the
convergence study. Since Lagged is the only approximation that is consistent
(Section \ref{sec:consistency}), we use it to compute the reference solution.

\subsection{Oscillating Conveyor Belt}

\begin{figure}[!h]
    \centering
    \includegraphics[width=0.8\columnwidth]{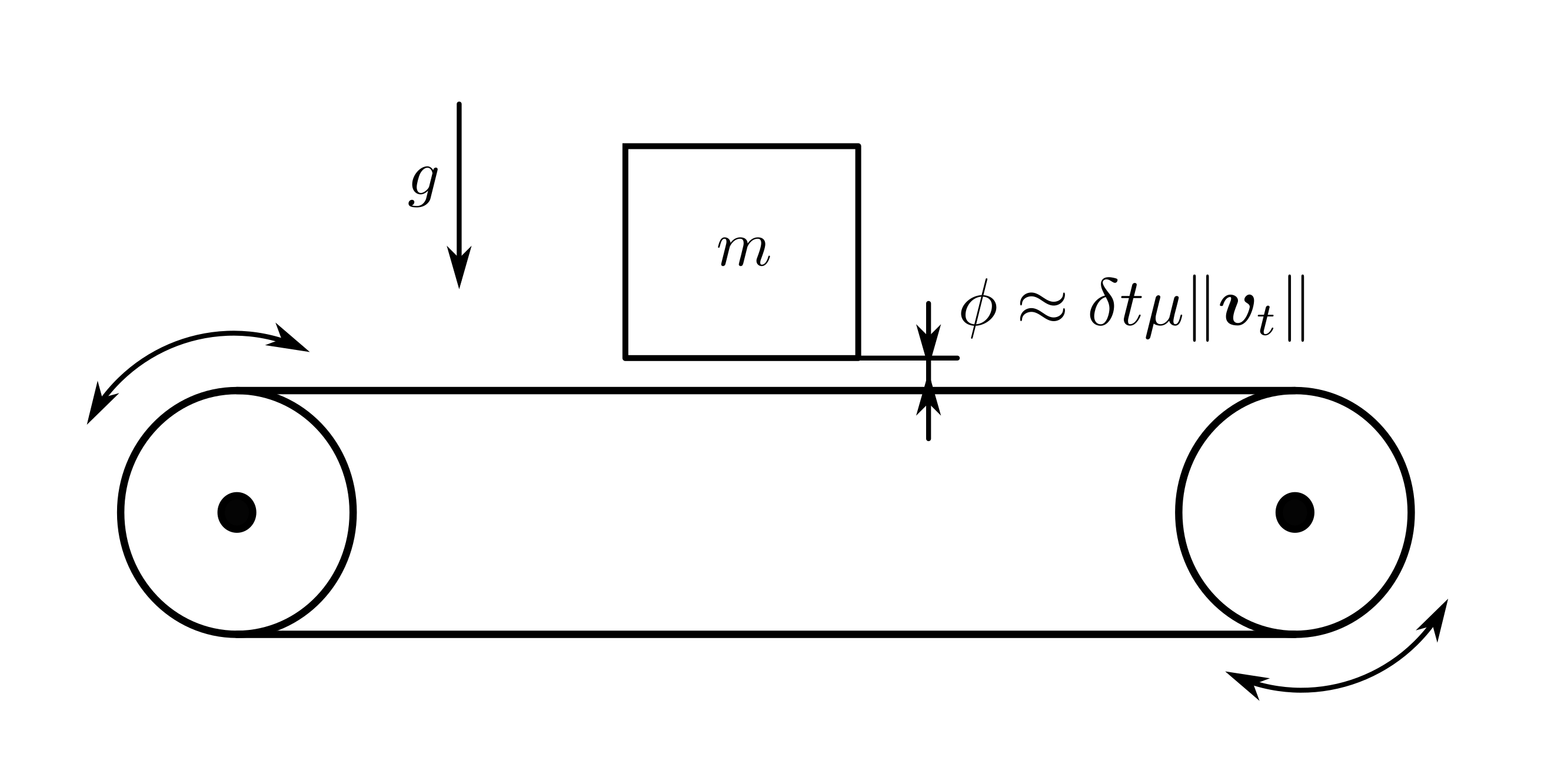}
    \caption{Oscillating Conveyor Belt. SAP and Similar introduce artificial
    \emph{gliding} during slip phases (Section \ref{sec:gliding_artifact}).}
    \label{fig:belt_shematic}
\end{figure}

This test illustrates artifacts in the strongly coupled SAP and Similar
approximations. A $1 \text{ kg}$ box with $5\text{ cm}$ sides is placed on a
conveyor belt oscillating at $1 \text{ Hz}$ with $0.2 \text{ m}$ amplitude (Fig.
\ref{fig:belt_shematic}). Friction is $\mu=0.7$. Even though dissipation models
are different, using $d=500 \text{ s/m}$ for Similar and Lagged and
$\tau_d=10^{-3} \text{ s}$ for SAP yields comparable dissipation.

\begin{figure}[!h]
    \centering
    \adjincludegraphics[width=0.49\columnwidth,trim={0 0 {0.05\width} 0},clip]{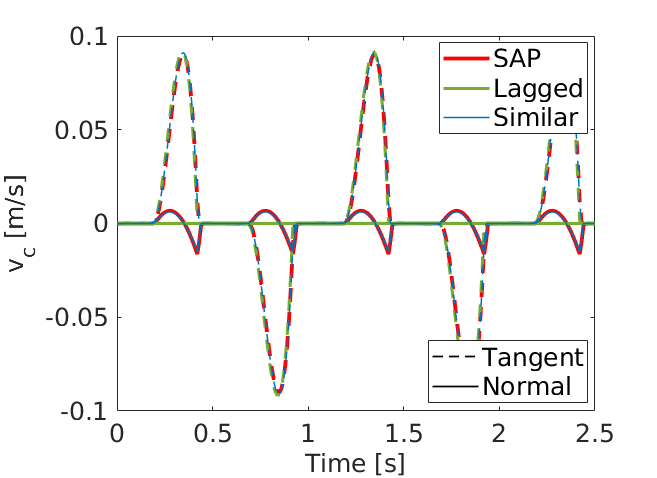}
    \adjincludegraphics[width=0.49\columnwidth,trim={0 0 {0.05\width} 0},clip]{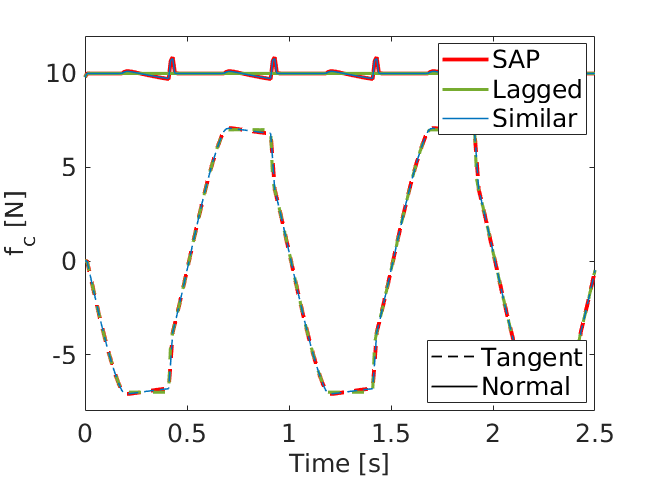}
    \caption{\label{fig:belt_contact_velocity_and_force} Contact velocity (left)
    and force (right). $\delta t=0.01\text{ s}$.}
\end{figure}

Figure \ref{fig:belt_contact_velocity_and_force} shows contact velocity and
force computed using $\delta t=0.01\text{ s}$. Contact between the box and belt
transitions back and forth between stiction and sliding. The Lagged
approximation predicts no vertical motion, with the normal force balancing the
box's weight as expected. However, SAP and Similar models show artifacts in the
normal direction during sliding, with non-zero normal velocity due to the
\emph{gliding} artifact, which disappears in stiction. Additionally, we observe
spurious transients in the normal force during sliding --- as slip speed changes
so does \emph{gliding}, causing vertical acceleration and thus normal force
fluctuations. Finally, SAP and Similar introduce normal force spikes
during the abrupt transition from sliding to stiction, when gliding vanishes
causing a sudden normal velocity change (Fig.
\ref{fig:belt_contact_velocity_and_force}).

\begin{figure}[!h]
    \centering
    \includegraphics[width=0.8\columnwidth]{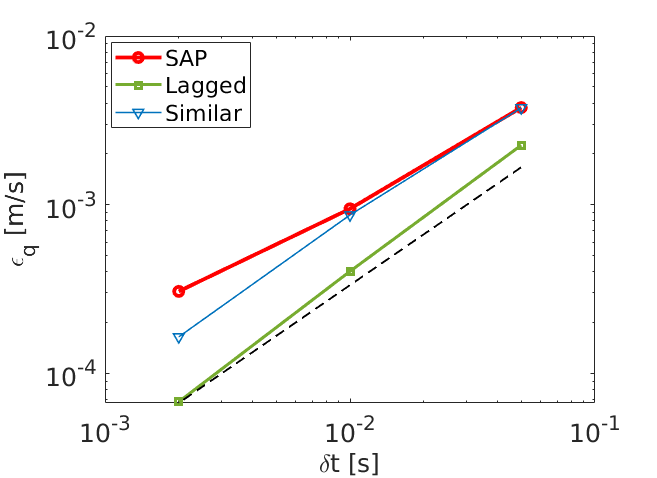}
    \caption{Convergence of the box trajectory with time step size. The dashed line is a reference for first order convergence.}
    \label{fig:bel_convergence_position}
\end{figure}

Figure \ref{fig:bel_convergence_position} shows a convergence study with step
sizes $\delta t \in \{2\times10^{-3}, 10^{-2}, 5\times10^{-2}\}$. Both Lagged
and Similar exhibit first order convergence, as expected, though the artifacts
introduced by Similar (Fig. \ref{fig:belt_contact_velocity_and_force}), cause
higher errors than Lagged. Finally, SAP's error plateaus at the smallest time
step due to model inconsistency, where the term $\tau_d\mu\Vert\vf{v}_t\Vert$ in
\eqref{eq:sap_gliding_offset} does not vanish as time step decreases.

\subsection{Falling Sphere}
\label{sec:falling_sphere}

The conveyor belt case from the previous section favors the Lagged model due to
the steady-state normal force, where the Lagged model is exact. Here, we
introduce a collision test to evaluate approximations under sudden contact
force changes.

\begin{figure}[!h]
    \centering
    \includegraphics[width=0.6\columnwidth]{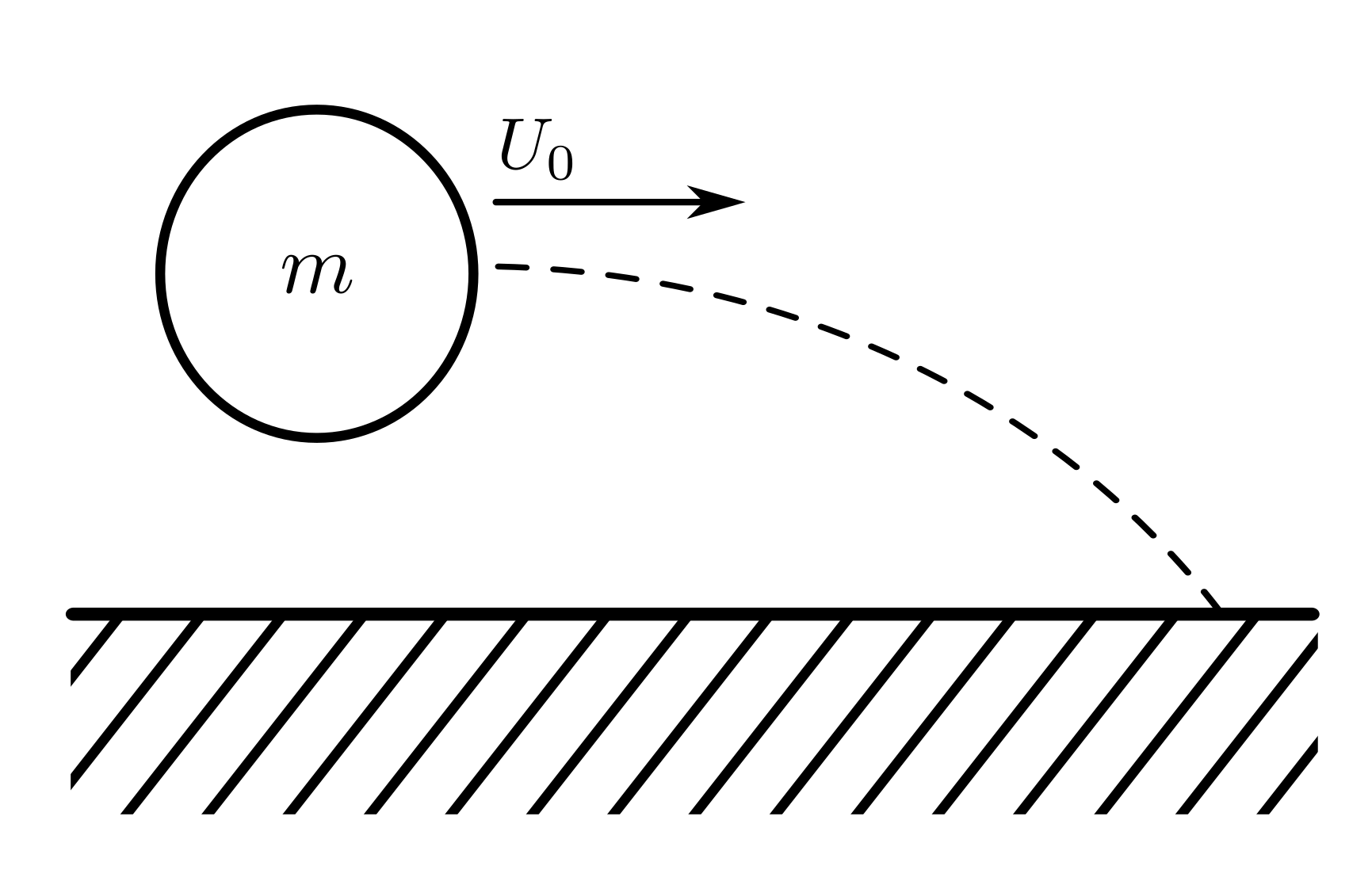}
    \caption{Falling sphere. After free fall, the sphere slides until friction with the ground establishes a rolling contact.}
    \label{fig:sphere_schematic}
\end{figure}

In this test, a $0.5\text{ kg}$ steel sphere $5\text{ cm}$ in diameter falls
from a height of $5\text{ cm}$ with an initial horizontal velocity $U_0=2\text{
m/s}$, see Fig. \ref{fig:sphere_schematic}. Friction with the ground is
$\mu=0.5$. Upon impact, the sphere slides, and then transitions to rolling
as friction induces angular momentum. After this transition, friction ceases to
dissipate energy. We model compliant contact with stiffness $k=10^{7}\text{
N/m}$ and dissipation constants $d=500\text{ s/m}$ and
$\tau_d=10^{-3}\text{ s}$.

\begin{figure}[!h]
    \centering
    \adjincludegraphics[height=0.38\columnwidth,trim={0 0 {0.05\width} 0},clip]{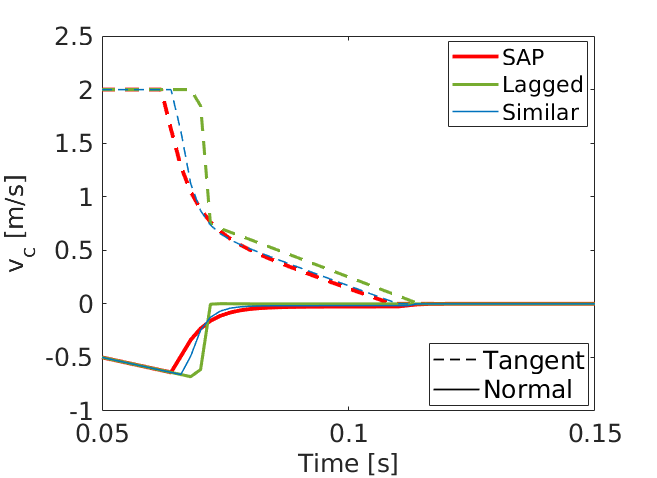}
    \adjincludegraphics[height=0.38\columnwidth,trim={0 0 {0.05\width} 0},clip]{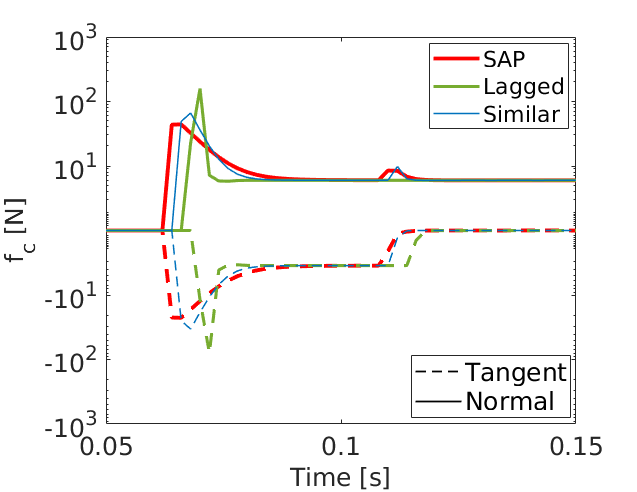}
    \caption{\label{fig:cylinder_contact_velocity_and_force} Contact velocity (left)
    and force (right) with $\delta t=2\times10^{-3}\text{ s}$.}
\end{figure}

Figure \ref{fig:cylinder_contact_velocity_and_force} shows contact velocity and
force computed with $\delta t=2\times10^{-3}\text{ s}$. In the force plots, both
SAP and Similar initiate contact earlier due to the \emph{action at a distance}
artifact in these convex models, with SAP engaging even earlier due to the
non-vanishing term $\tau_d\mu\Vert\vf{v}_t\Vert$ in
\eqref{eq:sap_gliding_offset} \reviewquestion{Q7}{(Section
\ref{sec:gliding_artifact})}. The sphere slides from initial contact until about
$t=0.07\text{ s}$, when it transitions to rolling. While Lagged brings normal
velocity to zero almost instantly, SAP and Similar models link normal velocity
to slip velocity, only reaching zero at stiction. During the sliding-to-stiction
transition, we observe a rapid normal force spike, as with the conveyor belt
problem. This artifact is absent with the Lagged approximation.

\begin{figure}[!h]
    \centering
    \includegraphics[width=0.8\columnwidth]{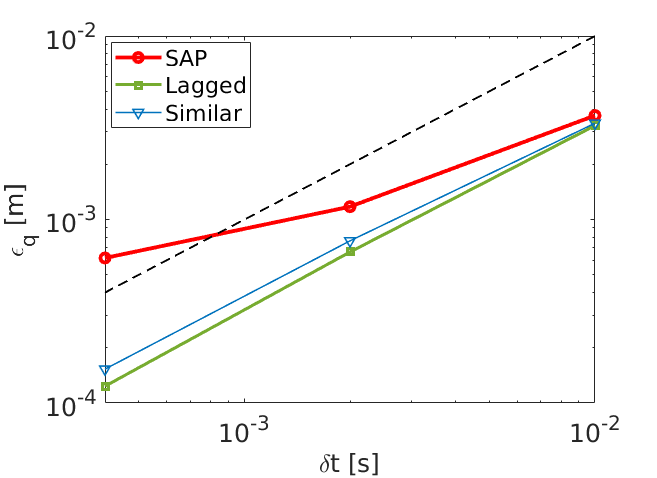}
    \caption{Convergence of the sphere trajectory with time step size. The dashed line is a reference for first order convergence.}
    \label{fig:cylinder_convergence_position}
\end{figure}

A convergence study with step sizes $\delta t \in \{4\times10^{-4},
2\times10^{-3}, 10^{-2}\}$ is shown in Fig.
\ref{fig:cylinder_convergence_position}. While all of these schemes are first
order, SAP exhibits a constant error at convergence due to the
$\tau_d\mu\Vert\vf{v}_t\Vert$ term in \eqref{eq:sap_gliding_offset}.

\subsection{Sliding Rod}
\label{sec:sliding_rod}

\begin{figure}[!h]
    \centering
    \includegraphics[width=0.55\columnwidth]{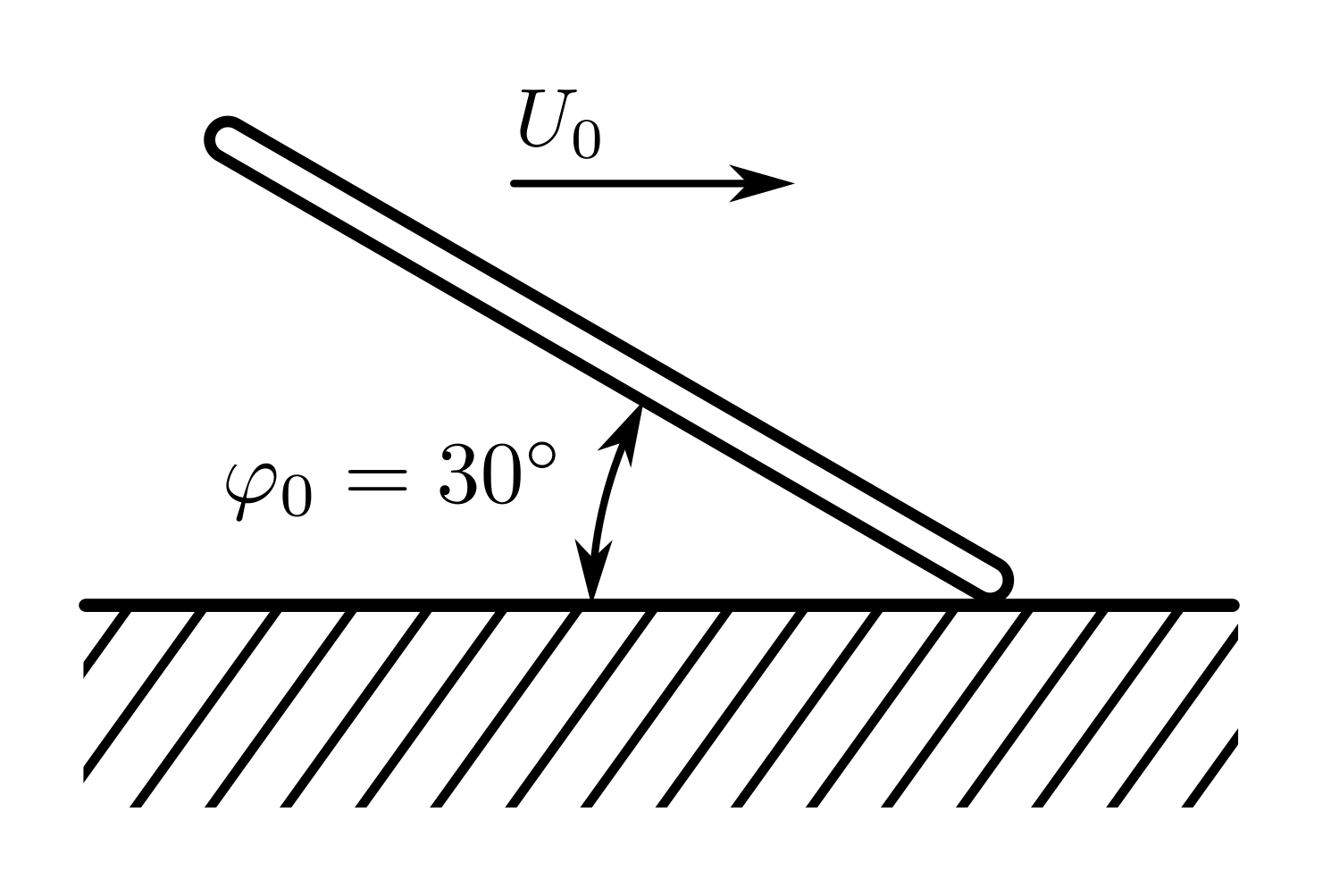}
    \caption{Sliding rod. Initially forming an angle $\varphi_0$ with the ground
    and with horizontal velocity $U_0$. Friction makes the rod rotate clockwise.
    The contact force increases until the rod jams into the ground, causing the
    rod to jump into the air.}
    \label{fig:sliding_rod_setup}
\end{figure}

This case is particularly interesting as it leads to impact without collision
\cite[\S 5.3]{bib:pfeiffer1996multibody}. A rod initially angled with the ground
makes single-point contact with a horizontal velocity (see Fig.
\ref{fig:sliding_rod_setup}). As it slides, friction rotates the rod into the
ground, increasing the normal force. Under specific conditions, both normal and
frictional forces intensify, potentially leading to a singularity in
acceleration-level formulations with Coulomb friction known as Painlev\'e's
paradox, where forces become infinite. This problem is resolved in the discrete
setting, where finite impulses and discrete velocity changes are allowed.
Physically, bodies aren't perfectly rigid --- they deform, vibrate, and may
even undergo plastic (permanent) deformations. Nonetheless, a rapidly increasing
contact force develops an impact that makes the rod jam into the ground and jump
into the air. The rod measures $0.5\text{ m}$ in length, $1\text{ cm}$ in
diameter, and has a mass of $0.3\text{ kg}$.

Analytical analysis \cite[\S 5.3]{bib:pfeiffer1996multibody} shows that the
singularity occurs when $\mu>4/3$ and the initial kinetic energy overcomes
potential energy as the rod's center of gravity rises and friction dissipates
energy. We set $\mu=2.3$, a $30^\circ$ initial angle (Fig.
\ref{fig:sliding_rod_setup}), and an initial horizontal speed $U_0=10\text{
m}/\text{s}$.

\begin{figure}[!h]
    \centering
    \adjincludegraphics[height=0.375\columnwidth,trim={0 0 {0.05\width} 0},clip]{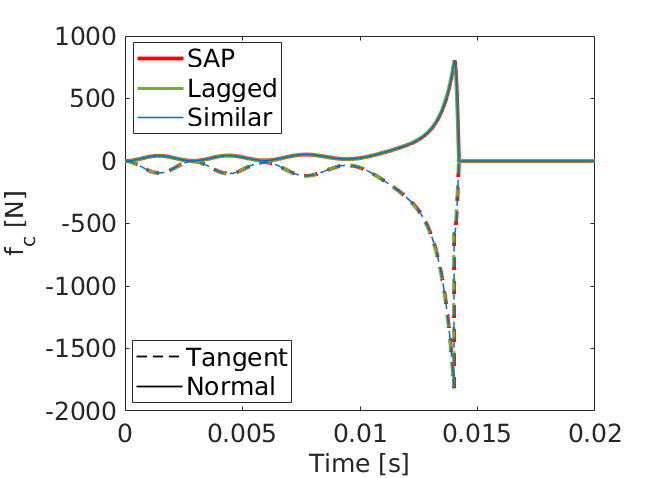}
    \adjincludegraphics[height=0.375\columnwidth,trim={0 0 {0.05\width} 0},clip]{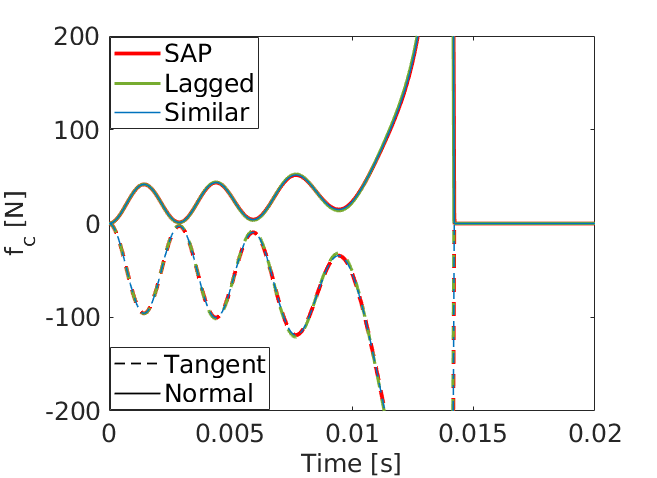}
    \caption{\label{fig:rod_contact_force_no_diss} Contact forces in the case with zero
    dissipation. The figure on the right shows a close-up near the impact.}
\end{figure}

Using a reference solution with a time step of $\delta t=10^{-7}\text{ s}$ and no
normal force dissipation, we observe that the rod rotates upward and jams into
the ground upon contact as expected. All three models yield similar results, as
the compliant model is identical in the absence of dissipation, differing only
in friction regularization. Pre-impact forces oscillate based on ground
compliance, and impact location is nearly identical across models (
Fig. \ref{fig:rod_contact_force_no_diss}) despite being very sensitive to model
parameters.

\begin{figure}[!h]
    \centering
    \adjincludegraphics[height=0.375\columnwidth,trim={0 0 {0.05\width} 0},clip]{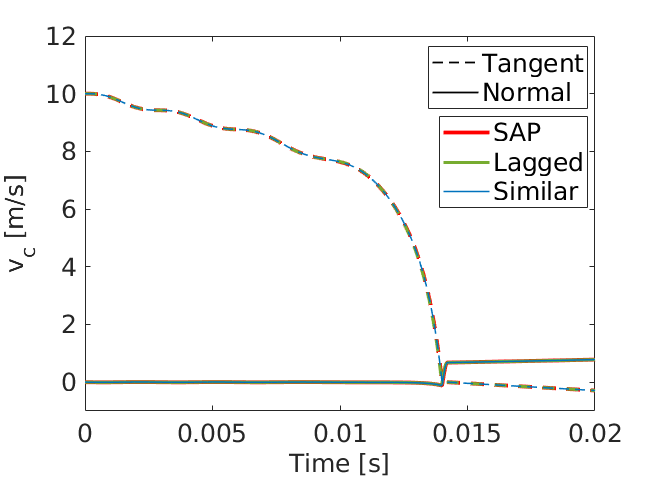}
    \adjincludegraphics[height=0.375\columnwidth,trim={0 0 {0.05\width} 0},clip]{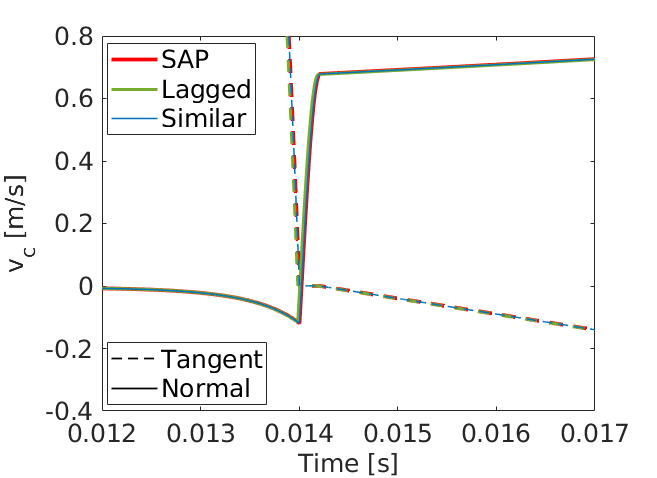}
    \caption{\label{fig:rod_contact_velocity_no_diss} Contact velocities in the case with zero
    dissipation. The figure on the right shows a close-up near the impact.}
\end{figure}

Figure \ref{fig:rod_contact_velocity_no_diss} shows the contact point moves
into the ground due to large contact forces until the tangential component of
the velocity goes to zero and the rod jams into the ground. During stiction
$\Vert\vf{v}_t\Vert<v_s$ for a finite period of about $0.2\text{ ms}$.

\begin{figure}[!h]
    \centering
    \adjincludegraphics[height=0.375\columnwidth,trim={0 0 {0.05\width} 0},clip]{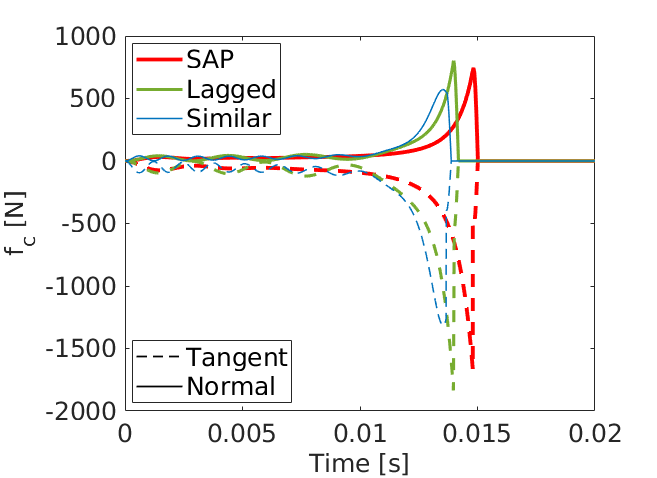}
    \adjincludegraphics[height=0.375\columnwidth,trim={0 0 {0.05\width} 0},clip]{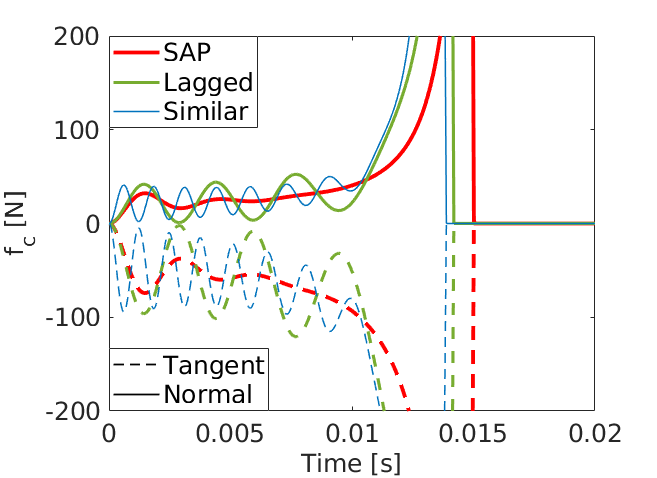}
    \caption{\label{fig:rod_contact_force} Contact forces in the case
    with dissipation. The figure on the right shows a close-up near the impact.}
\end{figure}

\begin{figure}[!h]
    \centering
    \adjincludegraphics[height=0.375\columnwidth,trim={0 0 {0.05\width} 0},clip]{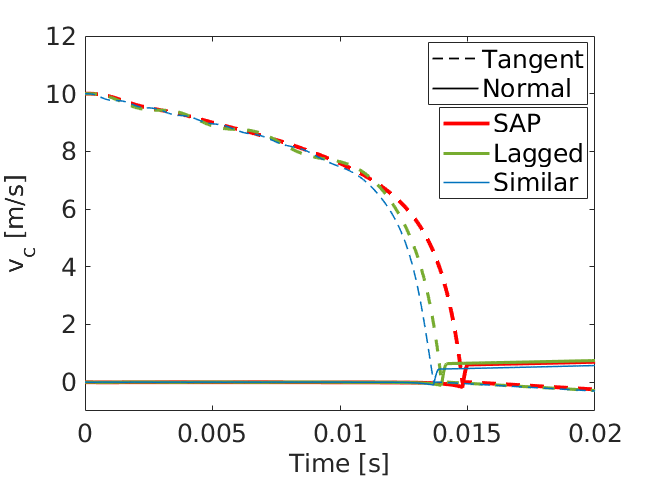}
    \adjincludegraphics[height=0.375\columnwidth,trim={0 0 {0.05\width} 0},clip]{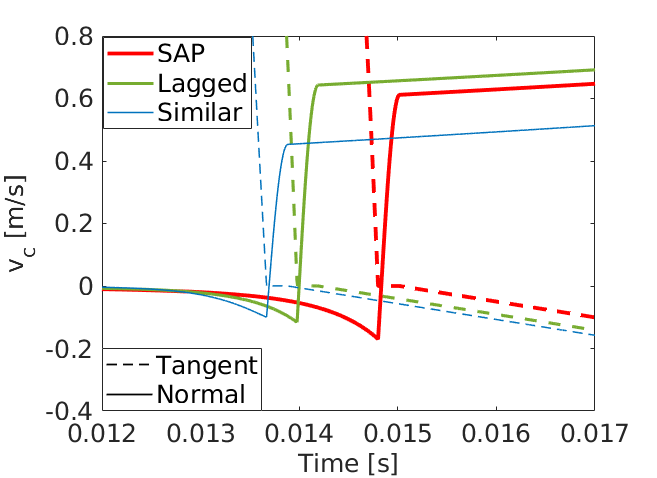}
    \caption{\label{fig:rod_contact_velocity} Contact velocities in the case
    with dissipation. The figure on the right shows a close-up near the impact.}
\end{figure}

We run the simulation with Hunt \& Crossley dissipation $d=0.2\text{ s}/\text{m}$
and relaxation time $\tau_d = 4.0\times{10}^{-6}\text{ s}$. These low
dissipation values have minimal effect on the time of impact for the Lagged model,
our reference solution, as seen in Figs. \ref{fig:rod_contact_force} and
\ref{fig:rod_contact_velocity}. However, Similar and SAP models predict shifted
impact times—earlier for Similar, later for SAP. Although their contact forces and
velocities differ significantly, we choose $\tau_d$ for SAP to match the shift in time
of impact observed in the Similar model, albeit in the opposite direction.
Compliance modulation in Similar becomes evident in Fig.
\ref{fig:rod_contact_force}, where we observe a frequency shift on the force
oscillations. This is caused by the larger effective stiffness of the model
during sliding, $k_\text{eff}=k\,(1+\mu\Vert\vf{v}_t\Vert\,d)$.

\begin{figure}[!h]
    \centering
    \adjincludegraphics[height=0.375\columnwidth,trim={0 0 {0.05\width} 0},clip]{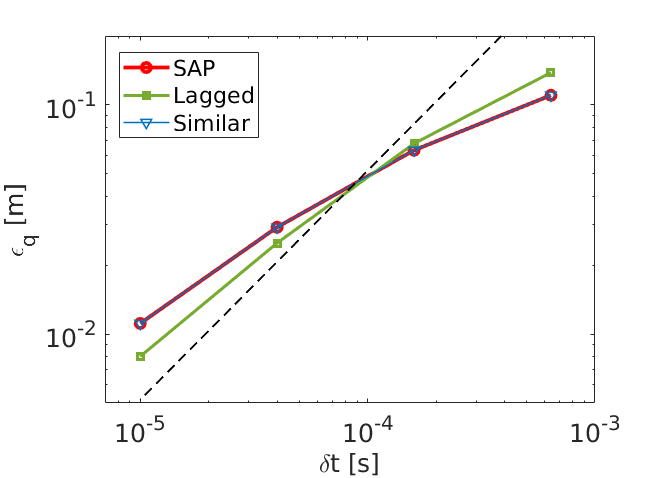}
    \adjincludegraphics[height=0.375\columnwidth,trim={0 0 {0.05\width} 0},clip]{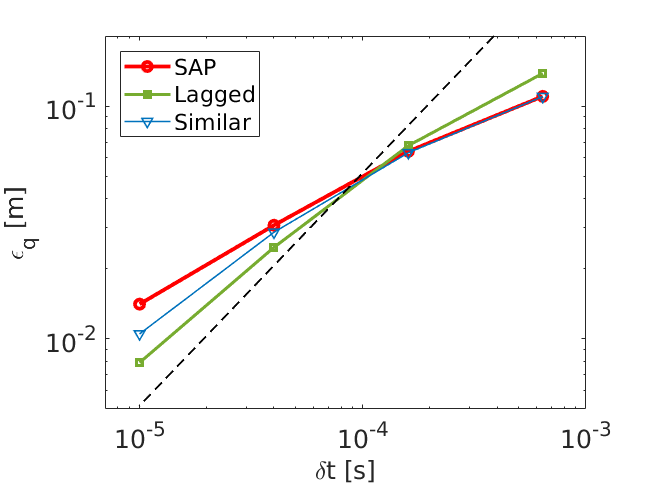}
    \caption{\label{fig:rod_convergence} Convergence in positions with
    time step. Case without dissipation (left) and with dissipation (right).}
\end{figure}

A convergence analysis with and without dissipation is shown in Fig.
\ref{fig:rod_convergence}, using time steps $\delta t=\{6.4\times 10^{-4},
1.6\times 10^{-4}, 4.0\times 10^{-5}, 1.0\times 10^{-5}\}\text{ s}$. Without
dissipation, SAP and Similar model solutions are indistinguishable. The Lagged
model exhibits the largest error at the largest time step. revealing a limitation:
its lagged normal force affects Coulomb friction modeling in rapidly changing
scenarios, even missing impacts when $\delta t > 10^{-3}\text{ s}$. However, it
achieves first-order convergence with smaller steps. Similar trends occur with
non-zero dissipation, though the curves diverge as time of impact predictions shift.

\section{Applications}
\label{sec:applications}

We simulate various robotics-relevant cases to evaluate model usefulness,
accuracy, numerical conditioning, and solver performance. In all simulations,
the solver iterates fully to convergence with no early termination, using a
relative tolerance $\varepsilon_r=10^{-5}$ \cite{bib:castro2022unconstrained}.

\subsection{Clutter}
\label{sec:clutter}

\begin{figure}[!h]
    \centering
    \adjincludegraphics[height=0.65\columnwidth,trim={0 0 {0.05\width} 0},clip]{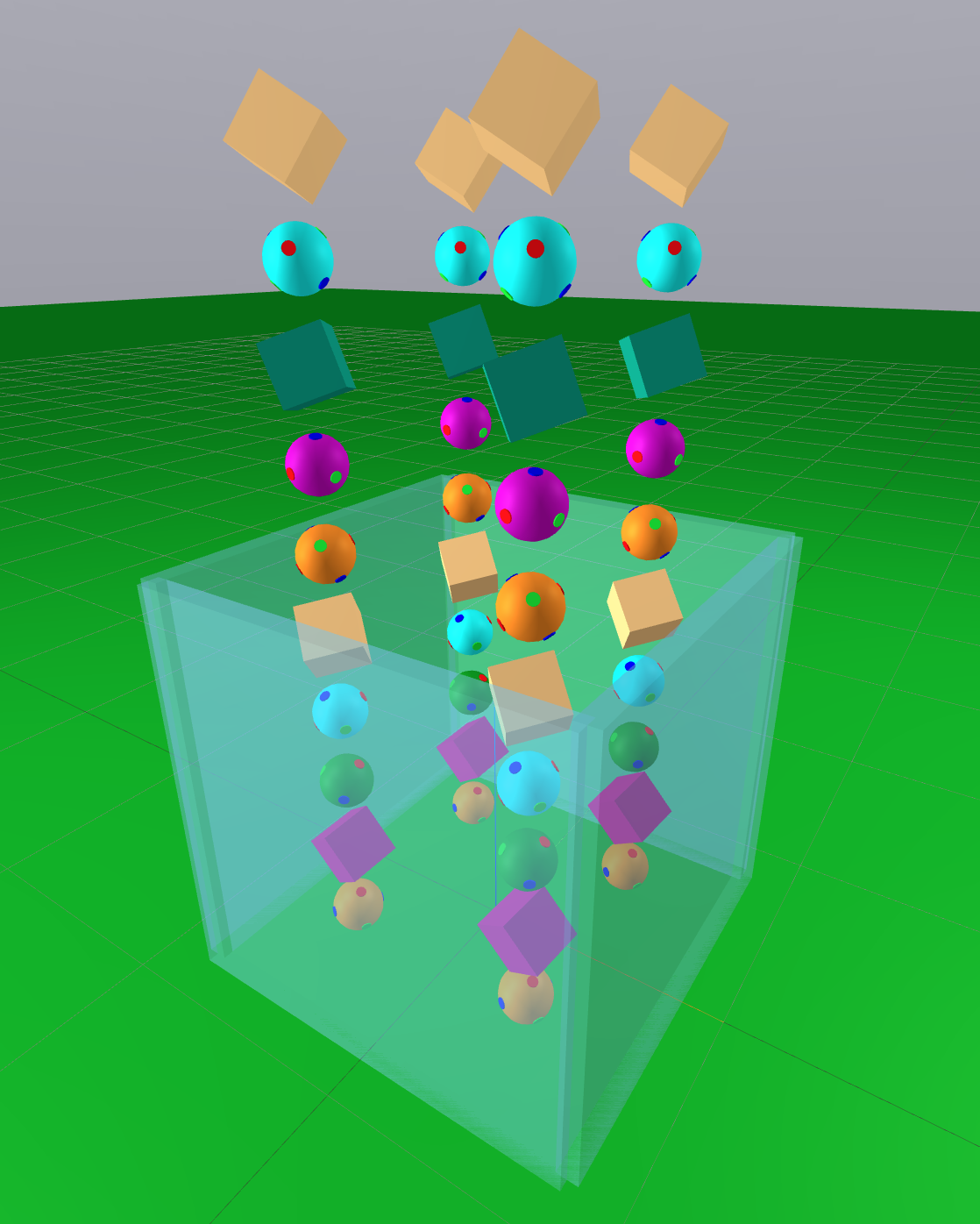}
    \adjincludegraphics[height=0.65\columnwidth,trim={0 0 {0.05\width} 0},clip]{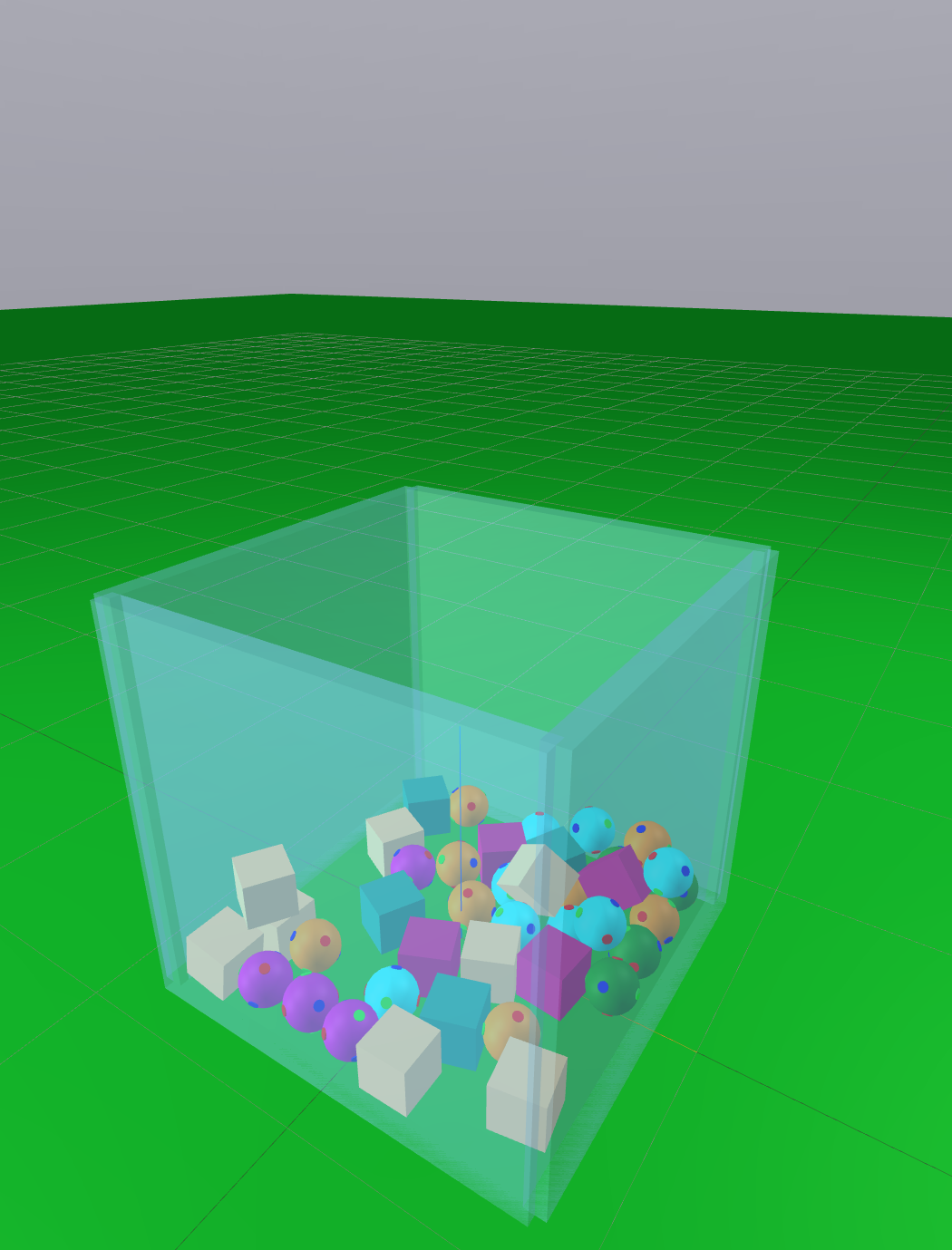}
    \caption{\label{fig:clutter_initial_final} Initial condition (left) and
    final steady state at $t=3$ seconds (right).}
\end{figure}

We reproduce the clutter experiment from \cite{bib:castro2022unconstrained} to
evaluate solver performance in cluttered environments, common in robotic
manipulation. We drop 40 objects, arranged in four columns of 10, into an $80
\times 80 \times 80\text{ cm}$ box, see Fig. \ref{fig:clutter_initial_final}.
Each column has a mix of $10\text{ cm}$ diameter spheres and $10\text{ cm}$
boxes, with masses calculated using water density: $0.524\text{ kg}$ for spheres
and $1.0\text{ kg}$ for boxes. A high stiffness of $k=10^{7}\text{ N}/\text{m}$
models steel, with Hunt \& Crossley dissipation set to $d=10\text{ s}/\text{m}$
and SAP dissipation at $\tau_d=10^{-4} \text{ s}$. Lagged and Similar models use
a stiction tolerance of $v_s = 10^{-4}\text{ m}/\text{s}$, while SAP and the
regularized Lagged model use $\sigma=10^{-3}$. All surfaces have a friction
coefficient $\mu=1.0$. We let objects fall and simulate for $3$ seconds.

\begin{figure}[!h]
    \centering
    \adjincludegraphics[height=0.375\columnwidth,trim={0 0 {0.05\width} 0},clip]{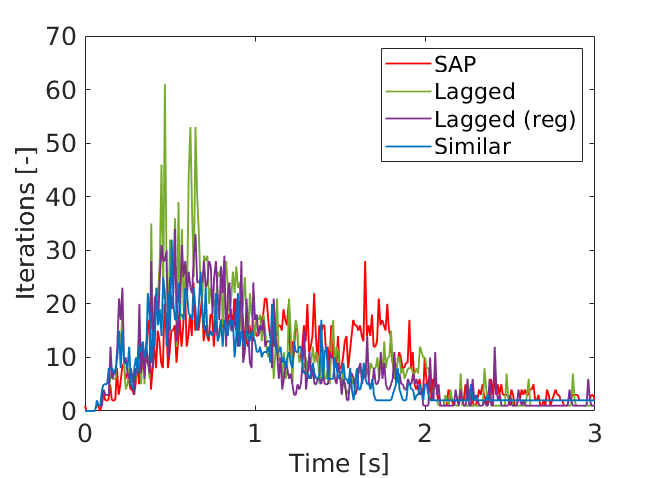}
    \adjincludegraphics[height=0.375\columnwidth,trim={0 0 {0.05\width} 0},clip]{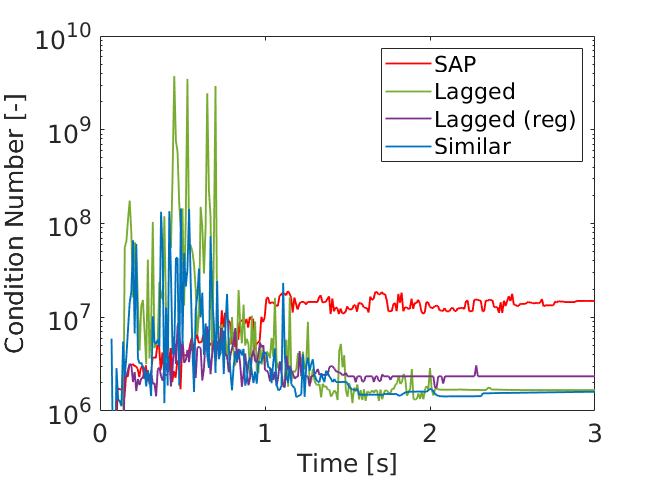}
    \caption{\label{fig:clutter_iters_cn} Iterations (left) and condition number
    (right) as a function of time. $\delta t=2\times10^{-3}\text{ s}$.}
\end{figure}

We analyze each method's performance with a time step $\delta
t=2\times10^{-3}\text{ s}$. Figure \ref{fig:clutter_iters_cn} shows Newton
solver iterations and Hessian condition numbers over time. An intense initial
transient occurs as objects collide upon falling, with impacts subsiding around
$t=2 \text{ s}$ as objects settle. During this early phase, the solver requires
more iterations, and condition numbers are higher. Notably, Lagged and Similar
approximations show higher condition numbers and iterations during the initial
transient, whereas the situation is reversed for SAP. This behavior relates to
the stiffness $G_t$ with regularized friction (Section
\ref{sec:impacts_and_conditioning}). Using the \emph{regularized} stiction
tolerance \eqref{eq:regularized_vs} in the Lagged approximation reduces
iterations and improves conditioning during the initial phase. Though not
included here, the same regularization can be used with Similar.

\begin{figure}[!h]
    \centering
    \includegraphics[width=0.8\columnwidth]{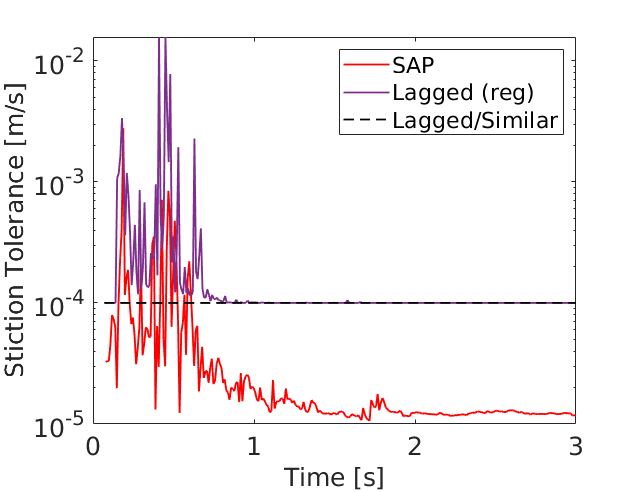}
    \caption{Effective stiction tolerance as a function of time. $\delta t=2\times10^{-3}\text{ s}$.}
    \label{fig:clutter_stiction_tolerance}
\end{figure}

Figure \ref{fig:clutter_stiction_tolerance} shows the effective stiction
tolerance. For Lagged and Similar approximations, stiction tolerance is fixed at
$v_s=10^{-4}\text{ m}/\text{s}$ (dashed black line). In contrast, SAP and the
regularized Lagged approximation have a tolerance that varies with normal impulse
(Section \ref{sec:impacts_and_conditioning}). This aligns with prior
observations: during the initial transient, Lagged and Similar models enforce a
tighter friction approximation. Past this transient, SAP unnecessarily solves a
much tighter approximation.

\begin{figure}[!h]
    \centering
    \adjincludegraphics[height=0.375\columnwidth,trim={0 0 {0.05\width} 0},clip]{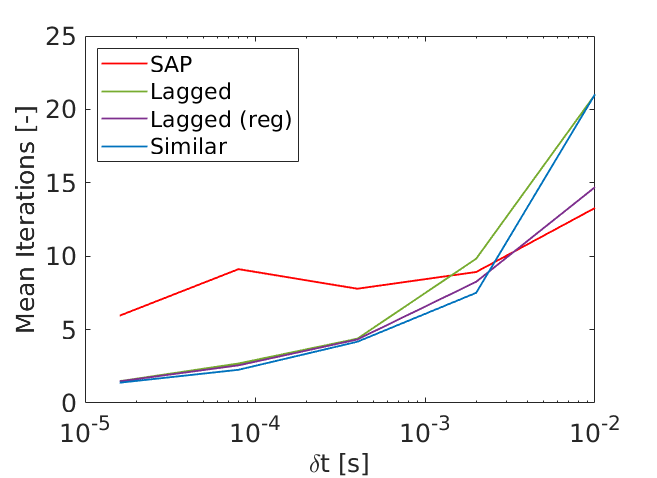}
    \adjincludegraphics[height=0.375\columnwidth,trim={0 0 {0.05\width} 0},clip]{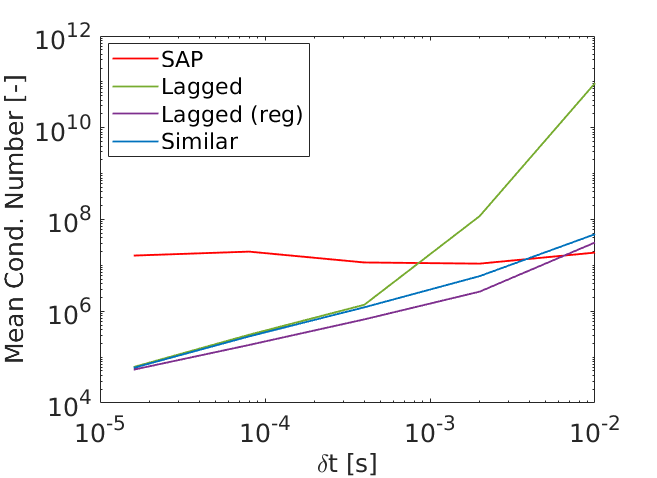}
    \caption{\label{fig:clutter_mean_iters_cn} Mean number of iterations (left) and
    condition number (right) per time step as a function of time step size.}
\end{figure}

An informative metric is the mean iteration count and condition number as a
function of time step, Fig. \ref{fig:clutter_mean_iters_cn}. For SAP,
the condition number remains almost unchanged across time step sizes since
$G_t=1/R_t$ (see Section \ref{sec:impacts_and_conditioning}) is constant. In contrast,
the conditioning of Lagged and Similar models improves as the time step decreases,
because $G_t$ is proportional to impulse and time step size. This reduction explains
why the number of iterations decreases for Lagged and Similar as the time step
decreases, while for SAP, it remains almost constant. Figure
\ref{fig:clutter_mean_iters_cn} also highlights the benefit of regularization
for Lagged at large time steps. At small time steps, $v_s$ dominates in
\eqref{eq:regularized_vs}, making Lagged with and without regularization perform
similarly.

\begin{figure}[!h]
    \centering
    \includegraphics[width=0.8\columnwidth]{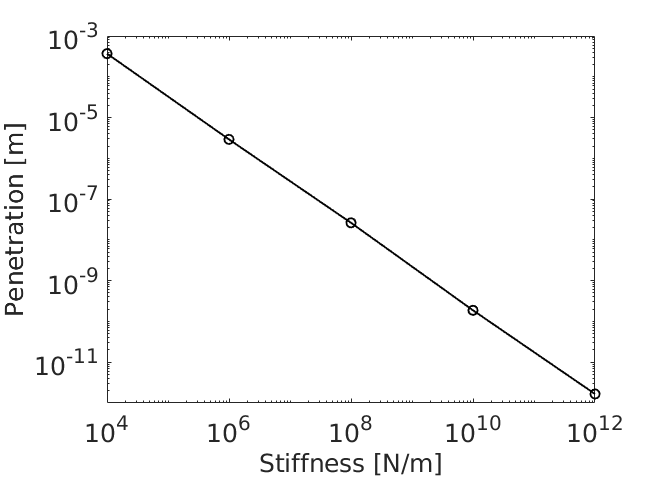}
    \caption{Mean penetration distance in steady state during the last $1.25\text{ s}$
    of the simulation.}
    \label{fig:clutter_penetration_study}
\end{figure}

We examine compliance as a method for approximating \emph{rigid} contact by
measuring mean penetration distance over the last $1.25$ seconds, when objects
settle at the bottom of the box. Figure \ref{fig:clutter_penetration_study}
shows steady-state penetration across stiffness values spanning eight orders of
magnitude. For reference, Hertz theory predicts a stiffness of $10^7\text{
N}/\text{m}$ for steel. We stress-test with stiffness up to five orders higher,
confirming the robustness of the convex formulation. For a time step $\delta
t=0.005\text{ s}$, SAP’s \emph{near-rigid} \cite{bib:castro2022unconstrained}
stiffness estimate is $10^6\text{ N}/\text{m}$, with penetration at only tenths
of microns. At an extreme, nonphysical $k=10^{13}\text{ N}/\text{m}$, the solver
fails due to round-off errors, while $10^5-10^6\text{ N}/\text{m}$ suffices for
approximating rigid contact in typical robotics applications.

\begin{figure}[!h]
    \centering
    \adjincludegraphics[height=0.375\columnwidth,trim={0 0 {0.05\width} 0},clip]{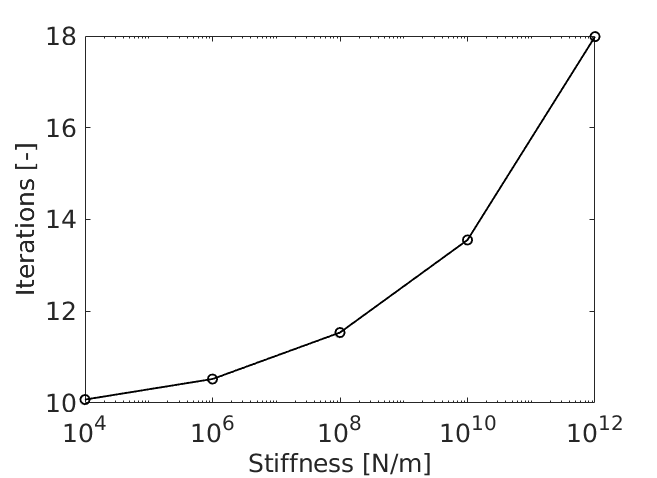}
    \adjincludegraphics[height=0.375\columnwidth,trim={0 0 {0.05\width} 0},clip]{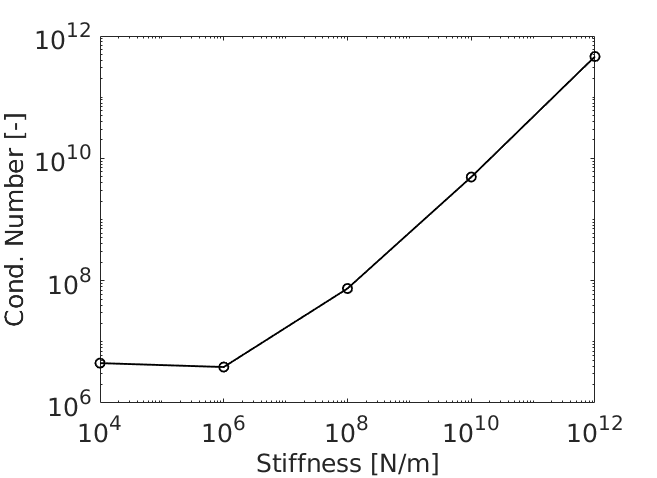}
    \caption{\label{fig:clutter_stiffness_study} Mean number of iterations (left) and
    condition number (right) per time step as a function of stiffness.}
\end{figure}

Finally, figure \ref{fig:clutter_stiffness_study} shows the effect of stiffness
on performance. As expected, higher stiffness values degrade conditioning and
ultimately impair performance. However, we note that the performance degradation
is minimal --- only within $20\%$, even for stiffness values as high as those of
steel.

\subsection{Grasp Stress Test}
\label{sec:franka_hand}

\begin{figure}[!h]
    \centering
    \adjincludegraphics[width=0.49\columnwidth]{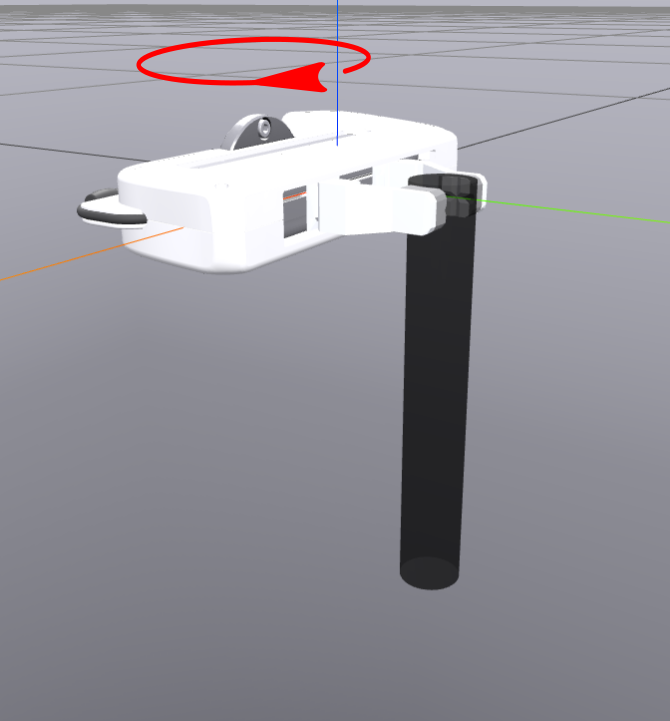}
    \adjincludegraphics[width=0.49\columnwidth]{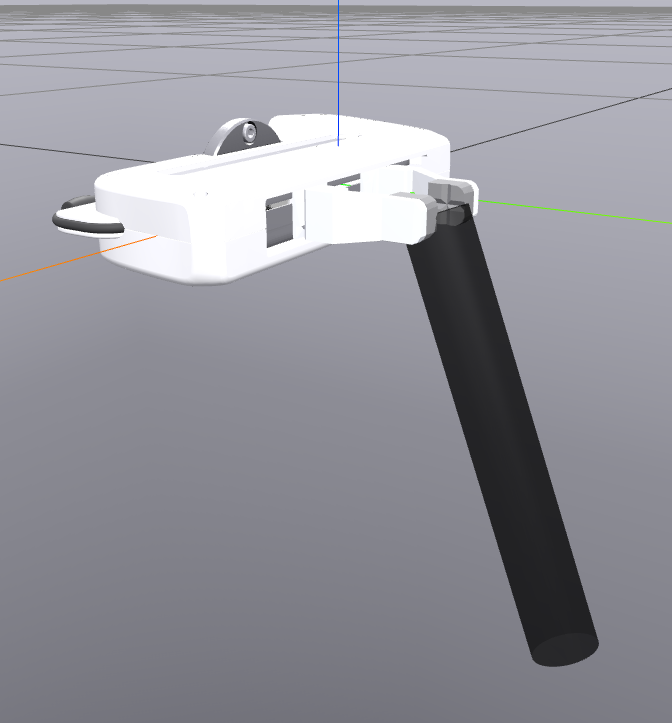}
    \caption{\label{fig:franka_snapshots} Franka hand holding a rod. Initial condition (left)
    and near grasp failure (right).}
\end{figure}

To evaluate grasp stability for Lagged and Similar approximations, we simulate a
Franka hand holding a bronze rod (density $8000~\text{kg}/\text{m}^3$) following
a 3~cm circular horizontal trajectory (Fig. \ref{fig:franka_snapshots}). At low
frequencies, the rod is secure, but near the rod's compound pendulum
frequency, it begins to rock, slide, and ultimately fall. We refer to this
time as \emph{time to failure} $T_f$.

We identify three error sources: $\mathcal{O}(\delta t)$ truncation error in the
symplectic Euler scheme \eqref{eq:scheme_momentum}, gliding error ($\mu\delta
t\Vert\vf{v}_t\Vert$), and compliance modulation. Tests are designed to evaluate
their significance and determine if weak coupling in the Lagged approximation
(Section \ref{sec:strong_coupling}) has a measurable impact compared to these
errors.

\begin{figure}[!h]
    \centering
    \includegraphics[width=0.8\columnwidth]{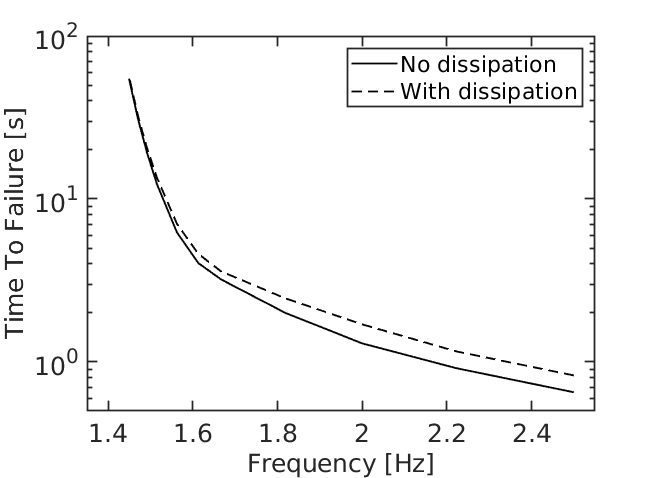}
    \caption{Time to Failure $T_f$ with and without Hunt \& Crossley dissipation. Reference solutions
    with $\delta t = 0.2\text{ ms}$.}
    \label{fig:ranka_reference_solutions}
\end{figure}

We compute reference solutions with Lagged (recall it is \emph{consistent}, Section
\ref{sec:consistency}) at a small time step of 0.2~ms to minimize truncation
errors (validated via a refinement study). To assess compliance modulation, we
compare solutions (Fig. \ref{fig:ranka_reference_solutions}) with dissipation
($d=50\text{ s/m}$) and without ($d=0$). Below 1.4 Hz, the grasp remains stable,
but $T_f$ decreases rapidly near the rod's compound pendulum frequency.
Dissipation slightly improves stability, increasing $T_f$ by up to 10\%.

\begin{figure}[!h]
    \centering
    \adjincludegraphics[height=0.38\columnwidth,trim={0 0 {0.05\width} 0},clip]{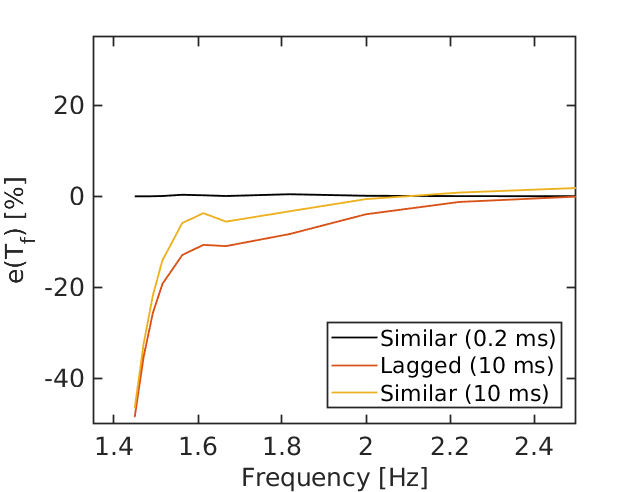}
    \adjincludegraphics[height=0.38\columnwidth,trim={0 0 {0.05\width} 0},clip]{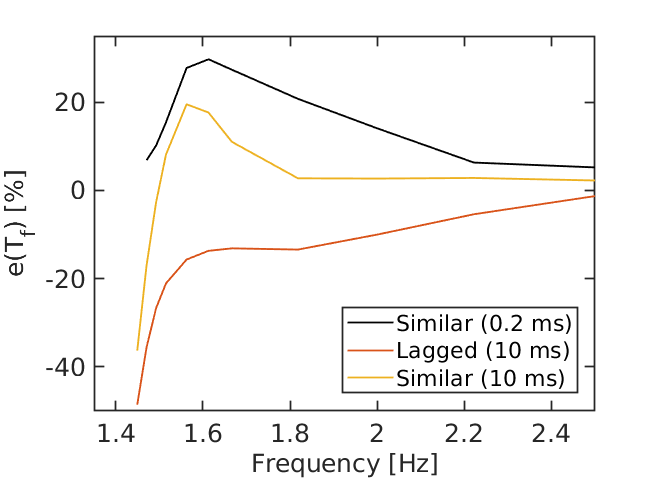}
    \caption{\label{fig:franka_time_to_failure_errors} Relative error in $T_f$ without
    (left) and with (right) dissipation. Positive values indicates $T_f$ is over predicted.}
\end{figure}

We compute the relative error in $T_f$ against the reference solutions (Fig.
\ref{fig:franka_time_to_failure_errors}), with positive values indicating
overestimation. At $\delta t=0.2\text{ ms}$ and zero dissipation, Lagged and
Similar solutions differ by less than 0.5\% (Fig.
\ref{fig:franka_time_to_failure_errors}, left), indicating that Similar's
\emph{gliding} contribution $\mu\delta t\Vert\vf{v}_t\Vert$ is negligible.

At large $\delta t=10\text{ ms}$ errors increase up to 50\% due to how
incredibly sensitive $T_f$ is. The goal however is to understand the relative
importance of each error contribution rather than a precise determination of
$T_f$. Without dissipation, truncation errors dominate, causing both
approximations to underpredict $T_f$. $T_f$ with Similar is slightly larger
compared to Lagged (error is less negative), as the gliding effect introduces
transient penetrations $\delta t\mu\Vert\vf{v}_t\Vert$ that increase the mean
grasp force.

From Fig. \ref{fig:franka_time_to_failure_errors} (left), Similar's gliding is
negligible at 0.2~ms. Thus, the 30\% error increase in Fig.
\ref{fig:franka_time_to_failure_errors} (right) is fully attributed to
compliance modulation, comparable to truncation error.

At $\delta t=10\text{ ms}$, Lagged's error remains nearly unchanged with or
without dissipation. Similar's error decreases due to the cancellation of two
effects: truncation errors causing $T_f$ underestimation (Fig.
\ref{fig:franka_time_to_failure_errors}, left) and compliance modulation leading
to overprediction (Fig. \ref{fig:franka_time_to_failure_errors}, right).

At large time steps for interactive simulation, both models exhibit similar
error magnitudes. Even in this highly dynamic case, Lagged's weak coupling
does not degrade grasp performance. Since Lagged eliminates \emph{gliding} and
\emph{compliance modulation}, we recommend it over Similar and SAP.

\subsection{BarrettHand}
\label{sec:barrett_hand}

\begin{figure}[!h]
    \centering
    \adjincludegraphics[width=0.49\columnwidth]{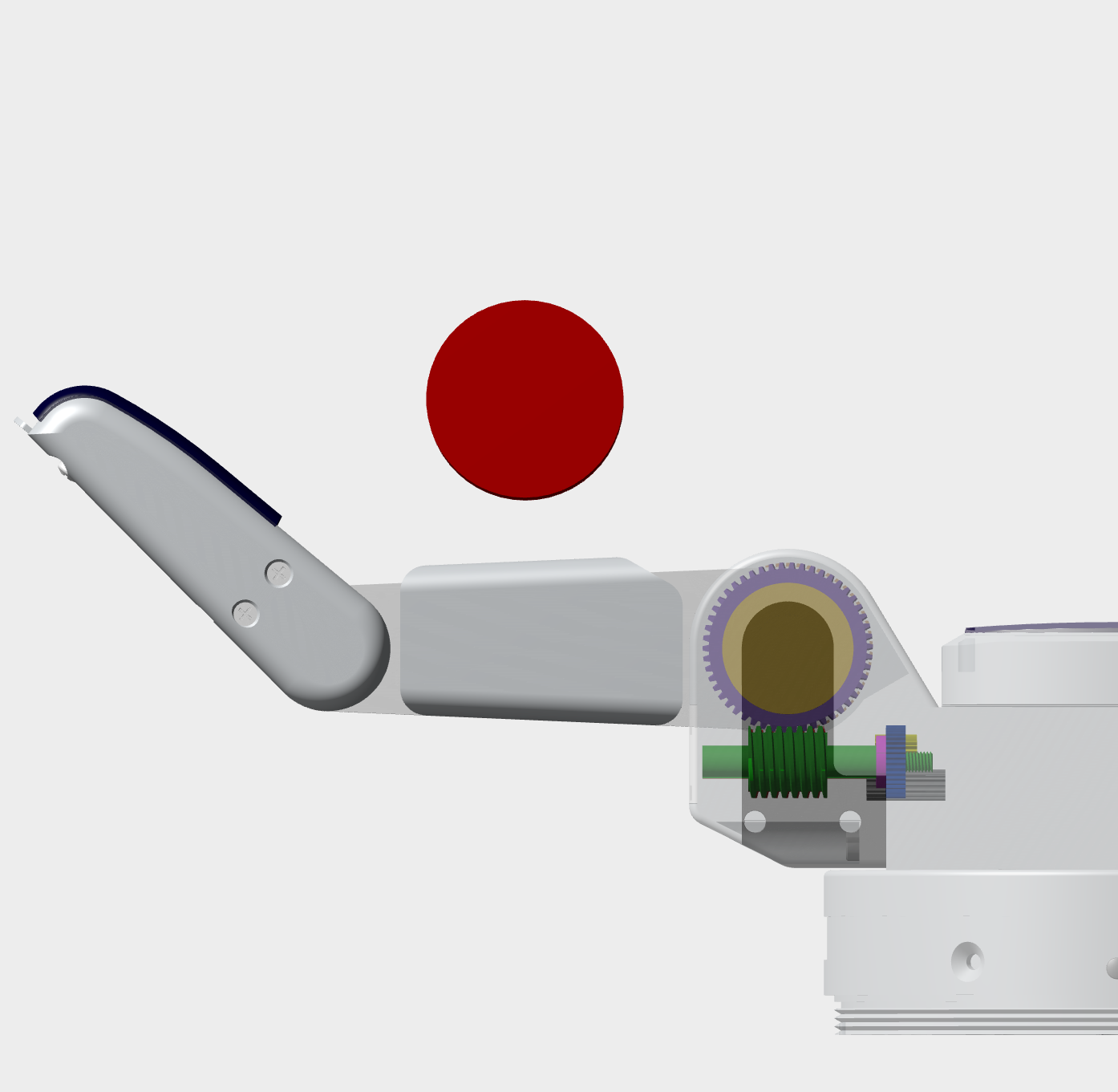}
    \adjincludegraphics[width=0.49\columnwidth]{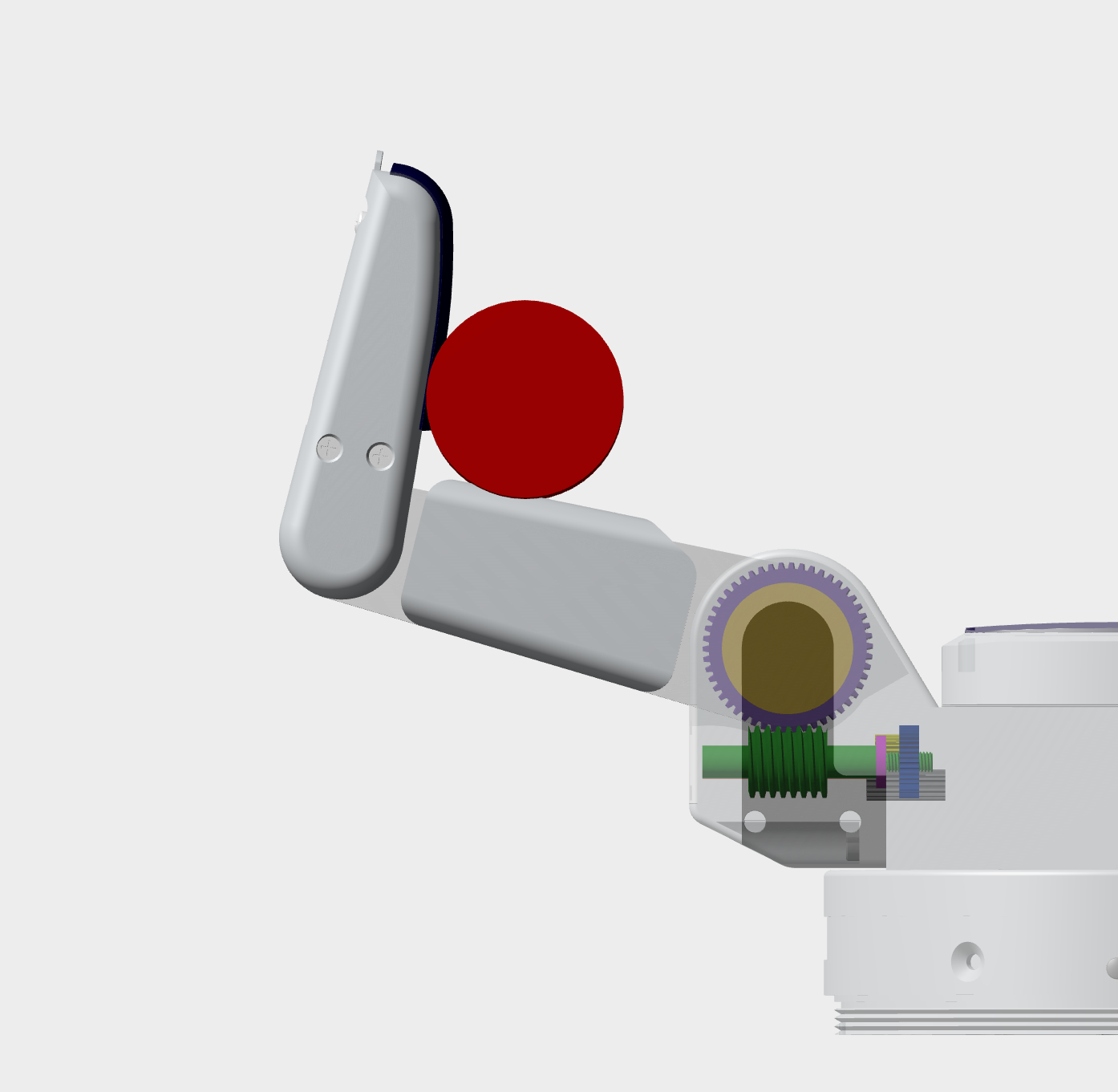}
    \adjincludegraphics[width=0.49\columnwidth, trim={0 0 0 {-0.15\columnwidth}}]{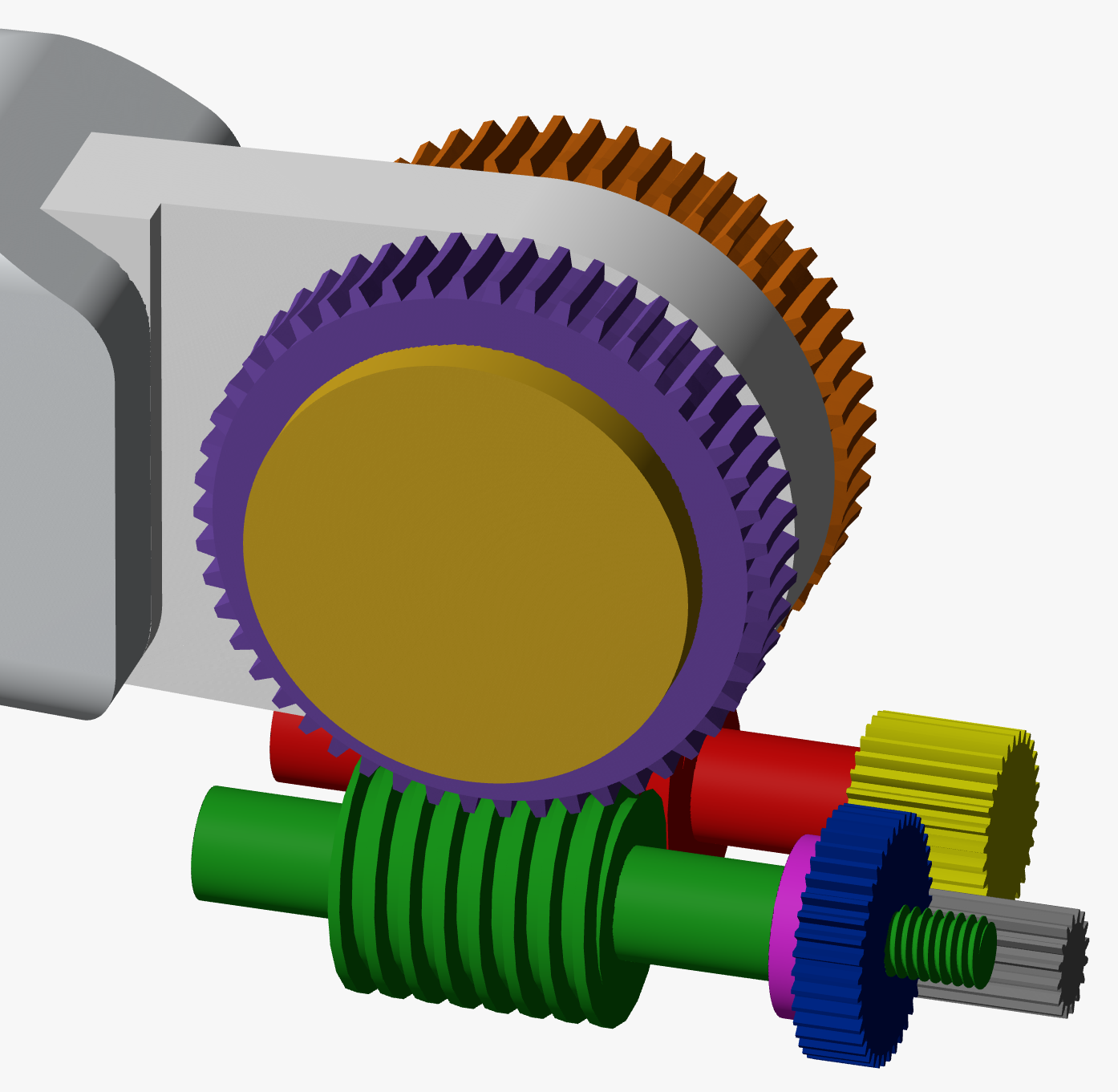}
    \adjincludegraphics[width=0.49\columnwidth, trim={0 0 0 {-0.15\columnwidth}}]{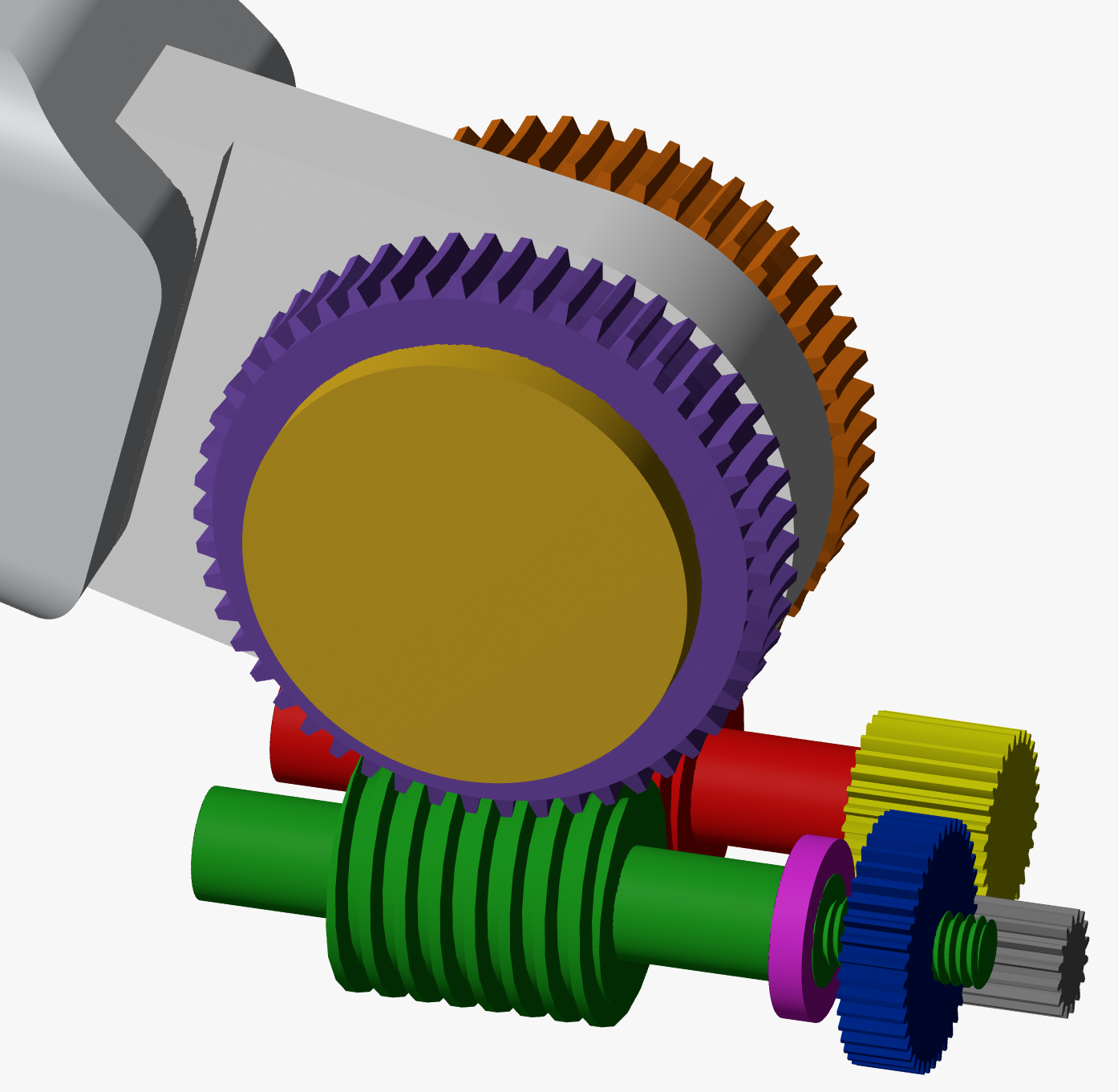}
    \caption{\label{fig:barrett_snapshots} Barrett hand finger with clutch
        engaged (left) and disengaged (right). Gears are color-coded in the proximal
        drivetrain as: proximal gear (blue), clutch (pink), proximal worm (green),
        proximal wheel (purple), and in the distal drivetrain as: distal gear
        (yellow), distal worm (red), distal wheel (orange).}
\end{figure}

We demonstrate our method's predictive capability by simulating the
BarrettHand's unique gearing mechanism~\cite{bib:barretthand}, which relies on
friction and stick-slip transitions. The proximal and distal links of a finger
are driven by a single motor through a coupled gear system. The proximal gear
(blue in Fig.~\ref{fig:barrett_snapshots}) rides freely on internal threads
along the shaft of the adjacent worm gear. Belleville washers at the end of the
threads act as a clutch for the proximal gear. When compressed by the proximal
gear (Fig.~\ref{fig:barrett_snapshots} left), the washers' stiction holds the
proximal gear stationary relative to the worm gear, transmitting torque from the
motor to the proximal link. When the proximal link reaches an external force
limit, the torque causes the proximal gear to slip and break away from the
washers (Fig.~\ref{fig:barrett_snapshots} right). At this point, the motor
torque transfers solely to the distal link. Stiction between the proximal wheel
(purple) and worm gear (green) makes the system non-backdrivable, holding the
proximal link in place.

We model a single BarrettHand finger along with its entire motor and gear
system. Our method captures the characteristic loading, driving, and breakaway
modes of the finger. Contact geometries are modeled to specification with
hydroelastic meshes~\cite{bib:elandt2019pressure} generated from CAD drawings.
The Belleville washers are modeled with a single compliant hydroelastic
cylinder (pink in Fig.~\ref{fig:barrett_snapshots}). The distal wheel (orange in
Fig.~\ref{fig:barrett_snapshots}) connects to the distal link via a pulley,
modeled with a holonomic constraint. The motor is driven by a PD controller with
effort limits to rotate at 200~rad/s. We set $\delta t = 0.5\text{ ms}$ to limit
tooth travel to 25\% of their width per time step. The friction coefficient
between clutch and proximal gear is 1.0 (effectively rough). For the worm gears,
the friction coefficient is estimated as $\mu=1.05\cdot\tan(\alpha)$, with
$\alpha$ the lead angle, to ensure non-backdrivability. All other surfaces are
frictionless. Overall, the system has 7 degrees of freedom. On average, there
are approximately 1100 contacts per time step.

\begin{figure}[!h]
    \centering
    \includegraphics[width=0.9\columnwidth]{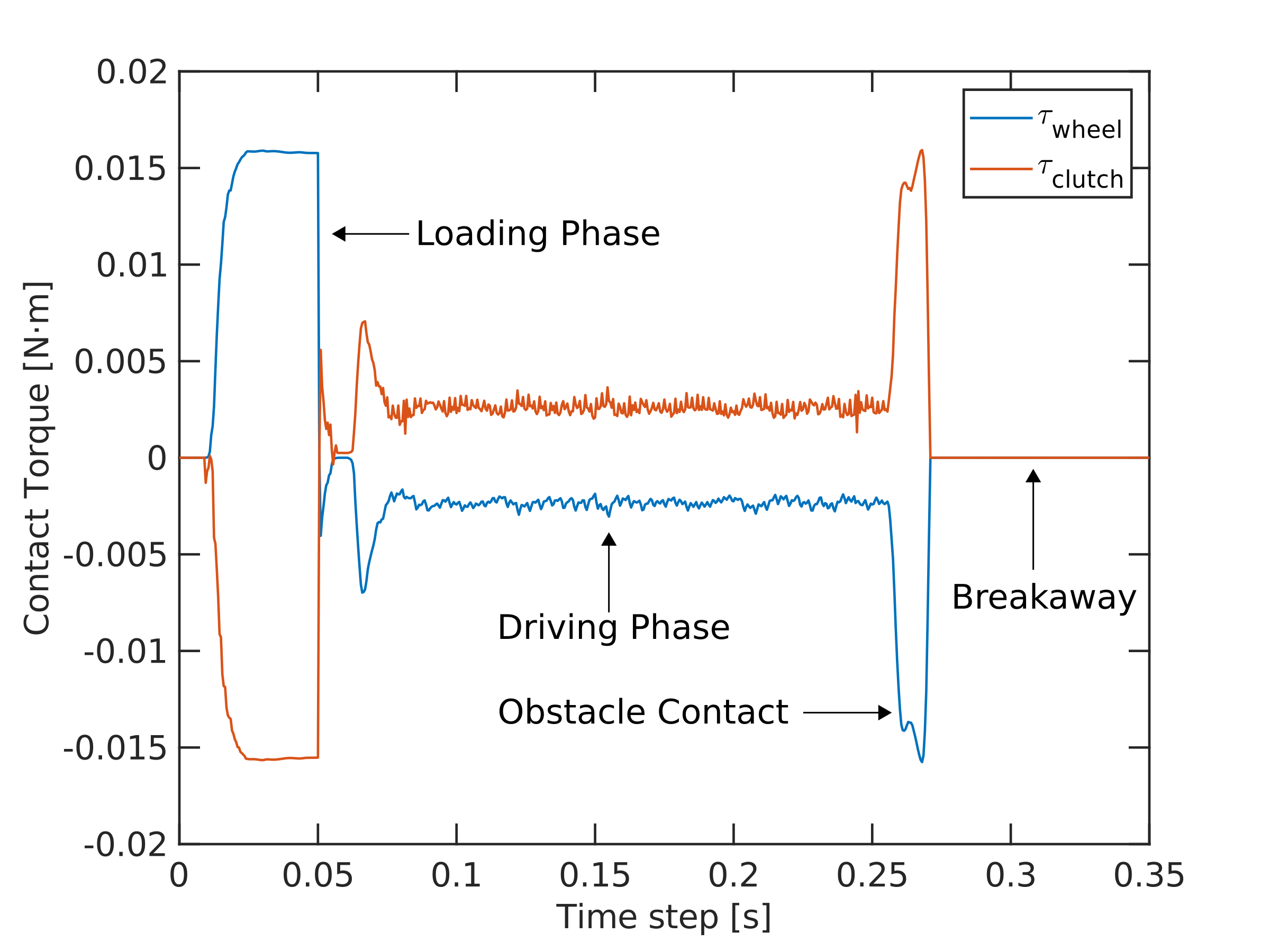}
    \caption{Torques on the proximal worm gear transmitted through contact from the proximal wheel and clutch.}
    \label{fig:barrett_torque}
\end{figure}

Figure \ref{fig:barrett_torque} shows the simulated contact torques on the
proximal worm gear during operation. In the initial configuration, the proximal
gear and clutch are not in contact. Thus, we drive the motor into a ``Loading
Phase'' such that the proximal gear is driven into contact with the clutch. Joint
limits and friction lock the proximal drivetrain while the proximal gear winds
down to meet the clutch until the motor reaches its effort limit (0.6~N$\cdot$cm). We
then reverse the motor direction at $t = 0.05\text{ s}$ to enter a ``Driving
Phase'' using a higher effort limit (0.66~N$\cdot$cm) --- proximal gear and clutch are
engaged as the resulting contact torques from the motor do not exceed the
magnitude of the loaded torque. We observe a characteristic ``hammering'' effect
in the contact torques as loads transfer from one tooth to the next. A frequency
analysis of the torques using a Fast Fourier Transform (FFT) verifies this,
containing harmonics consistent with the gear ratios, tooth widths, and angular
velocities. At $t = 0.26\text{ s}$ the proximal gear comes into contact with a
fixed obstacle. The clutch torque builds, resisted by contact between the now
fixed proximal wheel and worm gear, until it exceeds the limit reached during
the ``Loading Phase''. At this point, the proximal gear transitions into slip,
eventually winding up its threads until it is completely out of contact with the
clutch. The clutch disengages, and the net torque on the worm gear is zero; the
drivetrain is completely disconnected and stiction prevents the worm gear from
backdriving.

Finally, we point out that, with the Similar model, we observe a spurious
``locking'' of the proximal wheel's teeth as they glide within the threads of the
worm gear. This is similar to the phenomenon reported in Fig. 2 of
\cite{bib:horak2019}. Therefore, these simulation results use the Lagged model,
which does not introduce any of these artifacts during sliding.

\subsection{Trajectory Optimization}
\label{sec:trajopt}

\begin{figure}[!h]
    \centering
    \includegraphics[width=0.49\columnwidth]{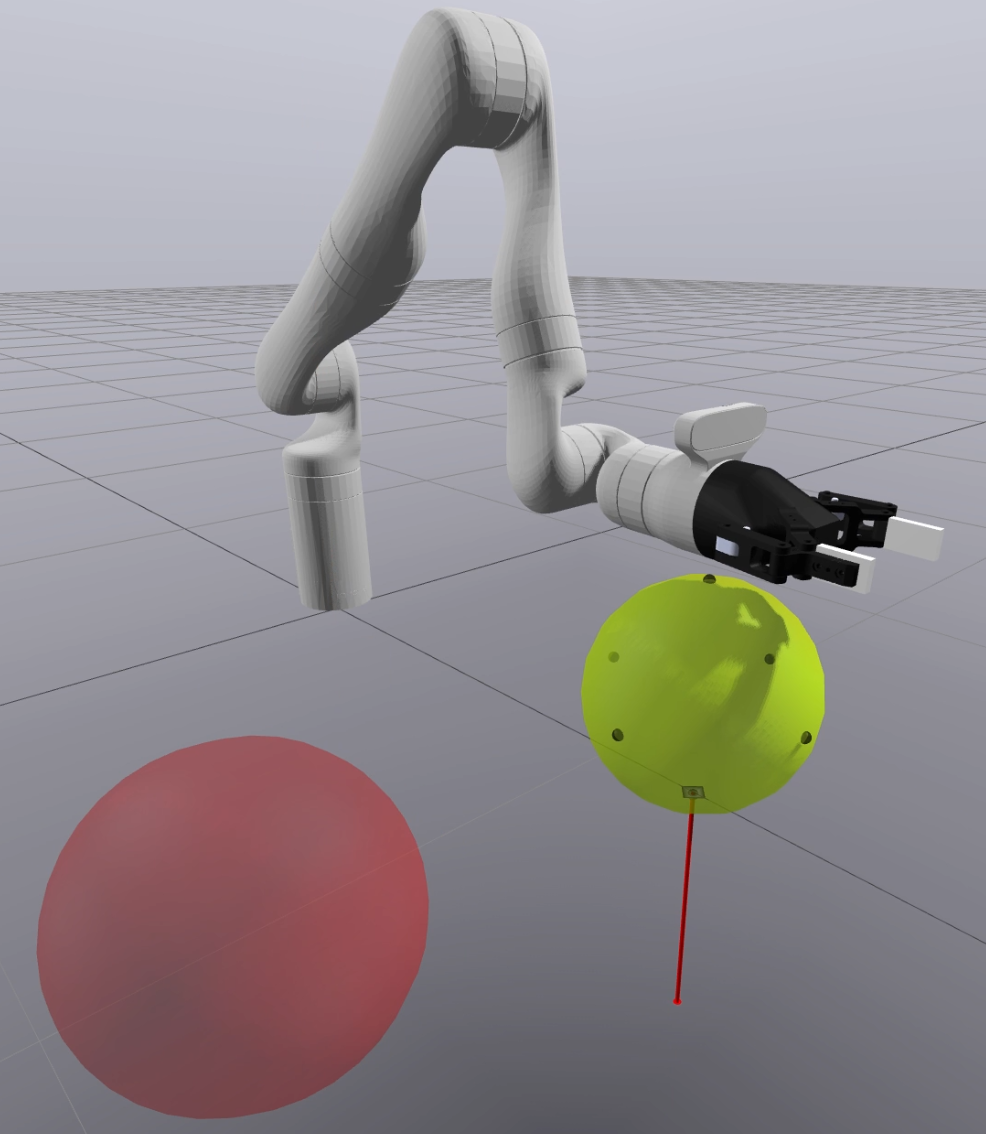}
    \includegraphics[width=0.49\columnwidth]{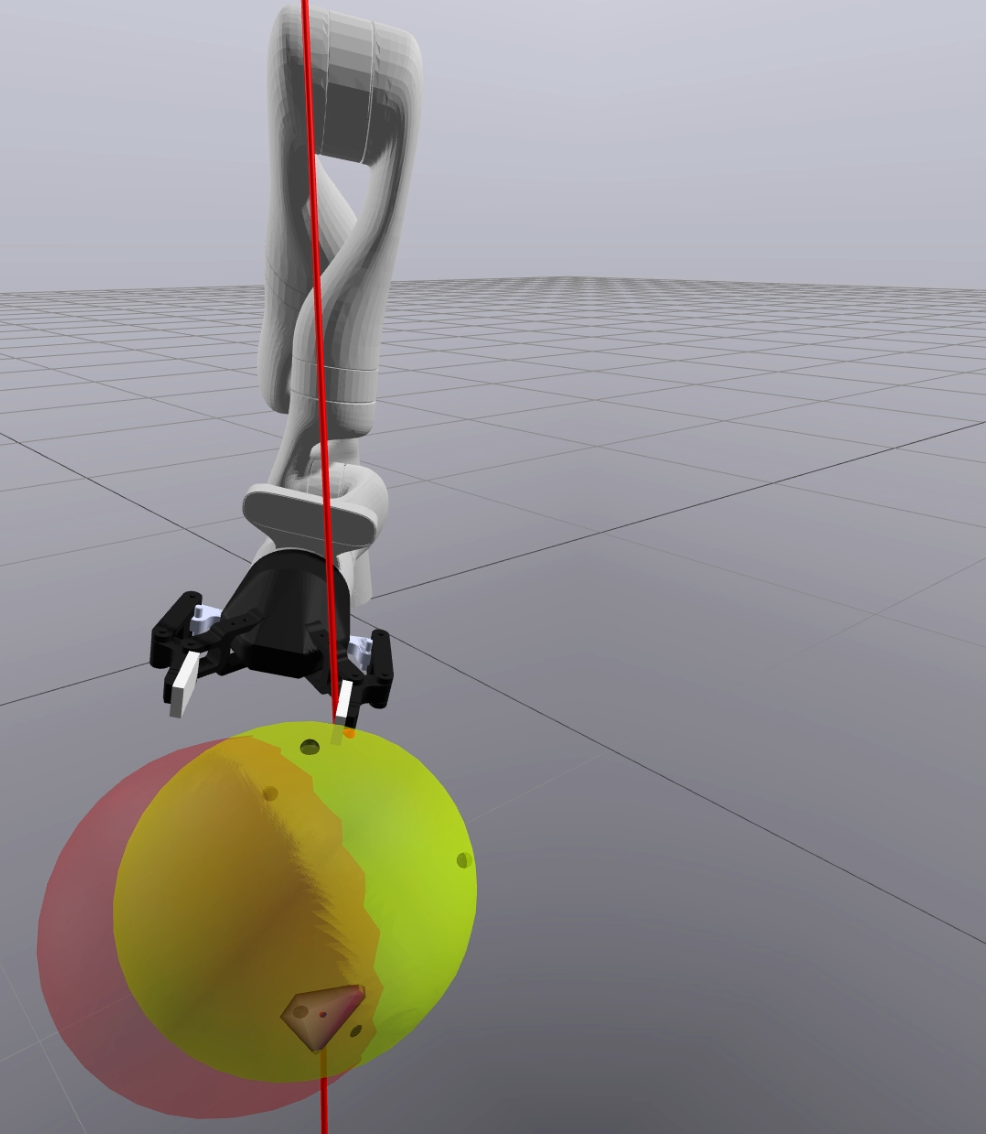}
    \caption{Keypoints of an optimal trajectory found by iLQR for the arm moving the ball to the target position (red).}
    \label{fig:kinova_gen3}
\end{figure}

We demonstrate end-to-end differentiation through contact using an
implementation of contact-implicit iLQR with Drake~\cite{bib:kurtz2022contact}.
To compute a local feedback controller and optimal trajectory, iLQR for a system
with discrete dynamics $x_{k+1} = f(x_k, u_k)$ must compute derivatives with
respect state $\partial f/\partial x_k$ and control inputs $\partial f/\partial
u_k$. In this demonstration, the task is for the Kinova Gen3 robot arm to move a
ball on a plane from a starting position to a target position
(Fig.~\ref{fig:kinova_gen3}). We use the exact model and environment setup
described in~\cite{bib:kurtz2022contact}, with all geometries modeled with
hydroelastic contact~\cite{bib:elandt2019pressure} and a time step of 10 ms over
a 0.5 s time horizon. The method is able to efficiently solve for an optimal
trajectory involving contact between the arm and the ball without explicitly
dictating a contact sequence.

\subsection{Deformable Bodies}
\label{sec:deformable}

\begin{figure}[!h]
    \centering
    \includegraphics[width=\columnwidth]{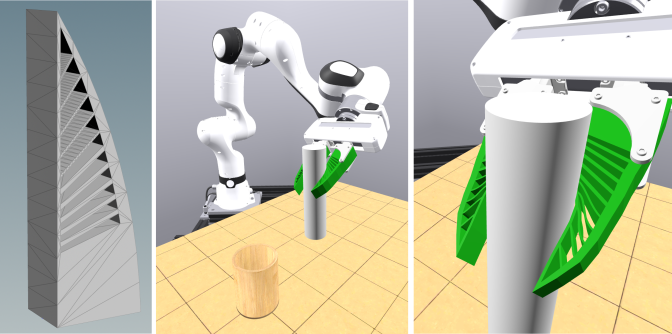}
    \caption{Simulation of deformable Finray grippers in a teleoperation task. (Left)
    the mesh used to model the FinRay gripper. (Center) the peg-in-hole teleoperation
    task. (Right) the characteristic caging deformation induced by frictional contact
    with the manipuland.}
    \label{fig:finray}
\end{figure}

Our method naturally extends to support frictional contact involving deformable
bodies. In this experiment, we simulate deformable FinRay gripper fingers
attached to a Panda arm in a peg-in-hole task (Fig.
\ref{fig:finray}). Each finger is discretized as a
tetrahedral mesh with 1,009 vertices and 2668 tetrahedra and simulated with the
linear corotational model described in \cite{bib:han2023}.
The fingers are attached to the Panda hand using holonomic constraints between
the mesh vertices the rigid hand. Material properties of the fingers include
a Young's modulus of $E=2.5 \times 10^6$ Pa, Poisson's ratio $\nu=0.49$, density
$\rho=1000$ kg/m$^3$, and Rayleigh stiffness damping coefficient $\zeta=0.01$ s.

The robot is teleoperated using a 6-DoF space mouse to control the end-effector
pose, while the arm's joint positions are solved through differential inverse
kinematics. The simulation employs the Lagged approximation with a 10 ms time
step. The robot is commanded to place a rigid cylinder into a rigid utensil
holder welded to the table. The average simulation time per time step is 24.5
ms, fast enough for interactive teleoperation, and the maximum and average
number of contact constraints are 114 and 73.9, respectively. 

\section{Conclusions}
\label{sec:conclusions}

We presented a novel theory for the convex approximation of contact. Our
mathematical framework establishes a family of convex approximations of
frictional contact, and we show that previous approaches
\cite{bib:anitescu2006,bib:todorov2011,bib:castro2022unconstrained} are members of
this family. This framework enables us to incorporate complex physics-based
models of contact, such as the Hunt \& Crossley \cite{bib:hunt_crossley} model,
within a convex formulation for the first time. These models, grounded
in physics and experimentally validated, have the potential to narrow the
\emph{sim2real} gap. Within this framework, we develop two convex approximations
of regularized friction: \emph{Similar} and \emph{Lagged}.

This work presents a thorough characterization of these approximations in
terms of consistency, the coupling between normal and tangential components, and
artifacts introduced by the convex approximation. While previous work has documented
\emph{gliding} during slip over a distance of $\delta t\mu\Vert\vf{v}_t\Vert$, we
identify previously unrecognized artifact characteristics of compliant contact
\cite{bib:todorov2011,bib:castro2022unconstrained}.
We validate these findings with a rich set of test cases designed to expose these
problems and gain insight into the new formulations. Moreover, our analysis
led to new understandings of the coupling between normal and frictional
components of the contact, allowing us to design a \emph{regularized} scheme that improves
numerical conditioning during difficult-to-resolve impact events. 

Our investigation concludes that, even though normal and friction forces are
weakly coupled, the Lagged approximation is well-suited for the modeling of most
robotic tasks. Moreover, the Lagged approximation completely eliminates
artifacts associated with previous convex approximations.

Our work is implemented in the open-source robotics toolkit Drake \cite{bib:drake}. 
We rigorously tested our implementation on various robotics-relevant problems, including
an iLQR application to highlight our differentiable pipeline and a deformable body
simulation to demonstrate compatibility with FEM-based methods.

One of the most significant limitations of our simulation pipeline is
related to \emph{tunneling} or \emph{passthrough} problems, where objects can
bypass each other without registering contact due to the nature of discrete contact detection.
This issue is particularly pronounced
with large time step sizes and thin objects. We are currently investigating a solution
based on \emph{speculative constraints} \cite{bib:catto} for hydroelastic contact
\cite{bib:elandt2019pressure,bib:masterjohn2021discrete} that we believe can
help mitigate this issue.

\appendices
\section{Soft Norm}
\label{sec:soft_norms}

We define the \emph{soft} norm of a vector $\mf{x}$ as
\begin{equation*}
    \Vert\mf{x}\Vert_s = \sqrt{\Vert\mf{x}\Vert^2+\varepsilon_x^2}-\varepsilon_x
    \label{eq:soft_norm_gradient},
\end{equation*}
where $\varepsilon_x>0$ has units of $\mf{x}$. Notice that $\Vert\mf{0}\Vert_s=
0$.

The gradient of the soft norm is
\begin{equation*}
    \frac{\partial\Vert\mf{x}\Vert_s}{\partial \mf{x}} = \frac{\mf{x}}{\Vert\mf{x}\Vert_s+\varepsilon_x} = \hat{\mf{x}}_s,
\end{equation*}
where we defined the \emph{soft unit vector} $\hat{\mf{x}}_s$.

The Hessian of the soft norm, the gradient of $\hat{\mf{x}}_s$, is
\begin{eqnarray*}
    \frac{\partial\hat{\mf{x}}_s}{\partial\mf{x}}=\frac{\partial^2\Vert\mf{x}\Vert_s}{\partial\mf{x}^2}&=&\frac{1}{\Vert\mf{x}\Vert_s+\varepsilon_x}\left(\mf{I}-\mf{P}(\hat{\mf{x}}_s)\right)\nonumber\\
    &=&\frac{\mf{P}^\perp(\hat{\mf{x}}_s)}{{\Vert\mf{x}\Vert_s+\varepsilon_x}}
\end{eqnarray*}
where the projection matrix is defined as
$\mf{P}(\hat{\mf{v}})=\hat{\mf{v}}\otimes\hat{\mf{v}}$. Note that
$\mf{P}(\hat{\mf{v}}) \succeq 0$ and $\mf{P}^\perp(\hat{\mf{v}}) \succeq 0$ for
all unit vectors $\hat{\mf{v}}\in\mathbb{R}^n$. Therefore, the soft norm is twice
differentiable with positive semi-definite Hessian, and thus it is convex.

These expressions for the gradient and Hessian of the norm of a vector are still
valid in the limit $\varepsilon_x\rightarrow 0$,  but they are not well-defined
at  $\mf{x}=\mf{0}$. However, the \emph{soft} versions have the nice
property that they are numerically well-behaved near and at $\mf{x}=\mf{0}$, and they are
continuously differentiable.

\section{Anisotropic Coulomb Friction}
\label{sec:anisotropic_friction}

When friction is
anisotropic, the friction force can have a component perpendicular to the slip
velocity and, even in the absence of external forces, objects follow curved
paths~\cite{bib:walker2019}. The ellipsoidal friction cone
$\mathcal{F}=\{[\vf{x}_t,
x_n]\in\mathbb{R}^3\,|\,\Vert\vf{\mu}^{-1}\vf{x}_t\Vert \le x_n\}$, with
$\vf{\mu}\succeq 0$ the friction tensor, is a popular approximation. $\vf{\mu}$
is diagonal when expressed in a frame aligned with its principal axes, and
$\vf{\mu} = \mu\mf{I}$ for isotropic friction.

We write an anisotropic model of Coulomb friction that satisfies
the principle of maximum dissipation
\begin{equation}
    \bgamma_{t}=\argmax_{\bxi\in\mathcal{F}}-\vf{v}_{t}\cdot\bxi.
    \label{eq:maximum_dissipation_principle}
\end{equation}

We can solve this problem analytically by a change of variables that maps the
ellipsoidal section of the friction cone into a circular section. The result is
\begin{equation}
    \bgamma_{t}=-\vf{\mu}\gamma_n\hat{\vf{t}}(\vf{\mu}\vf{v}_{t}),
    \label{eq:anisotropic_friction_model}
\end{equation}
where we define
$\hat{\vf{t}}(\vf{\mu}\vf{v}_{t})=\vf{\mu}\vf{v}_{t}/\Vert\vf{\mu}\vf{v}_{t}\Vert$.
This friction model opposes slip when tensor $\vf{\mu}$ is isotropic, while it
introduces a component perpendicular to the line of motion, as
experimentally confirmed in~\cite{bib:walker2019}.

As with~\eqref{eq:generic_friction_model}, we write a generic form of
this model as
\begin{eqnarray}
    \bgamma_{t}=g(\Vert\tilde{\vf{v}}_t\Vert, v_n)\,\tilde{\vf{t}},
    \label{eq:generic_anisotropic_model}
\end{eqnarray}
where we defined the \emph{tilde} quantities as
$\tilde{\vf{v}}_t=\vf{\mu}\,\vf{v}_{t}$ and
$\tilde{\vf{t}}=\vf{\mu}\,\hat{\vf{t}}(\tilde{\vf{v}}_t)$, consistent with
\eqref{eq:anisotropic_friction_model}. We verify that
\begin{equation*}
    \frac{\partial\bgamma_t}{\partial\mf{v}_t}=\frac{\partial g}{\partial \Vert\tilde{\vf{v}}_t\Vert}\tilde{\mf{P}}+g\frac{\tilde{\mf{P}}^\perp}{\Vert\tilde{\vf{v}}_t\Vert},
\end{equation*}
with $\tilde{\mf{P}}=\mf{P}(\tilde{\vf{t}})$ and $\tilde{\mf{P}}^\perp =
\vf{\mu}\mf{P}^\perp(\hat{\vf{t}}(\tilde{\vf{v}}_t))\vf{\mu}=\vf{\mu}^2-\tilde{\mf{P}}$.
Therefore, $\partial\bgamma_t/\partial\mf{v}_t$ for~\eqref{eq:generic_anisotropic_model}
is symmetric and condition~\eqref{eq:curl_normal} is met.

Using these \emph{tilde} variables, we define a potential for Lagged
\begin{equation*}
    \ell_t(\vf{v}_t) = \gamma_{n0}\,\varepsilon_s\,F(\Vert\tilde{\vf{v}}_t\Vert/\varepsilon_s),
\end{equation*}
and a potential for Similar is obtained by updating \eqref{eq:similar_grouping}
to $z=v_n-\varepsilon_sF(\Vert\tilde{\vf{v}}_t\Vert/\varepsilon_s)$.

\section*{Acknowledgment}
The authors would like to thank the Dynamics \& Simulation and Large Behavior Model 
teams at TRI for their continuous patience and support. We
thank William Townsend and Barrett for providing us with technical
specifications for the BarrettHand as well as advice and discussions on our
modeling approach. We thank Cody Simpson for modeling the deformable FinRay gripper.

\ifCLASSOPTIONcaptionsoff
  \newpage
\fi

\bibliographystyle{./IEEEtran/IEEEtran}
\bibliography{contact_fields}


\newpage



\vfill

\end{document}